\documentclass{IEEEtran}

\usepackage[utf8]{inputenc} %
\usepackage[T1]{fontenc}    %
\usepackage[
    hidelinks,
    pdftitle={Investigating the Impact of Action Representations in Policy Gradient Algorithms},
    pdfauthor={Jan Schneider, Pierre Schumacher, Daniel Häufle, Bernhard Schölkopf, Dieter Büchler}
]{hyperref}       %
\usepackage{url}            %
\usepackage{booktabs}       %
\usepackage{amsfonts}       %
\usepackage{nicefrac}       %
\usepackage{microtype}      %
\usepackage[labelformat=simple]{subcaption}
\usepackage{graphicx}
\usepackage{placeins}
\usepackage{amsmath}
\usepackage[capitalize,noabbrev]{cleveref}
\usepackage[sorting=none,maxbibnames=99,mincitenames=1,maxcitenames=2,style=ieee]{biblatex}
\usepackage{float}

\newcounter{para}
\newcommand{\numericparagraph}[1]{\par\refstepcounter{para}\textit{\thepara)\space #1}:}

\newcommand{\pos}{\mathbf{q}}
\newcommand{\vel}{\dot{\mathbf{q}}}
\newcommand{\state}{\mathbf{s}}
\newcommand{\act}[1]{\mathbf{a}_{#1}}

\newcounter{subsubfigure}
\newenvironment{subsubfigures}{%
  \refstepcounter{subfigure}%
  }{%
  \addtocounter{subfigure}{-1}%
  \setcounter{subsubfigure}{0}%
}
\newenvironment{subsubfigure}[1]{%
  \begin{subfigure}[b]{#1}%
    \refstepcounter{subsubfigure}%
    \addtocounter{subfigure}{-1}%
  }{%
  \end{subfigure}%
}

\newlength{\plotfigsep}
\newlength{\plotfigwidth}
\newlength{\surfacefigwidth}
\newlength{\surfacefigwidthappendix}
\title{Investigating the Impact of Action Representations \\ in Policy Gradient Algorithms}

\author{
    \IEEEauthorblockN{
        Jan Schneider\IEEEauthorrefmark{1}, Pierre Schumacher\IEEEauthorrefmark{1}\IEEEauthorrefmark{2}, Daniel Häufle\IEEEauthorrefmark{2}\IEEEauthorrefmark{3}, Bernhard Schölkopf\IEEEauthorrefmark{1}, Dieter Büchler\IEEEauthorrefmark{1}
    } \\[0.5\baselineskip]
    \IEEEauthorblockA{\IEEEauthorrefmark{1} Max Planck Institute for Intelligent Systems, Tübingen, Germany} \\
    \IEEEauthorblockA{\IEEEauthorrefmark{2} Hertie-Institute for Clinical Brain Research, Tübingen, Germany} \\
    \IEEEauthorblockA{\IEEEauthorrefmark{3} Institute for Modelling and Simulation of Biomechanical Systems, University of Stuttgart, Germany}
}

\begin{document}

\maketitle

\begin{abstract}
Reinforcement learning~(RL) is a versatile framework for learning to solve complex real-world tasks.
However, influences on the learning performance of RL algorithms are often poorly understood in practice.
We discuss different analysis techniques and assess their effectiveness for investigating the impact of action representations in RL.
Our experiments demonstrate that the action representation can significantly influence the learning performance on popular RL benchmark tasks.
The analysis results indicate that some of the performance differences can be attributed to changes in the complexity of the optimization landscape.
Finally, we discuss open challenges of analysis techniques for RL algorithms. 
\end{abstract}
\section{Introduction}
\label{sec:introduction}

Reinforcement learning~(RL) holds great potential for learning complex behaviors for solving real-world robotics tasks~\cite{kober2013reinforcement}.
However, the performance of RL methods often hinges on seemingly small implementation details~\cite{engstrom2020implementation} and design decisions in the task formulation~\cite{reda2020learning}.
These findings indicate that the inner workings of popular RL algorithms are often poorly understood.
Recent works tackle this issue by introducing analysis techniques for investigating different aspects of the learning process~\cite{ilyas2020closer,sullivan2022cliff}.
In this work, we evaluate and discuss the effectiveness of these analysis methods for investigating the impact of action representations in RL.

Recent results suggest that action representations can have a significant influence on the learning performance of RL algorithms~\cite{peng2017learning,varin2019comparison,bogdanovic2020learning,wochner2022learning}.
However, these works consider only a narrow set of custom-designed tasks.
Furthermore, the reasons for the performance differences are still unclear.
Understanding the causes of these results would enable designing action spaces that are particularly suited for RL, potentially improving learning efficiency and robustness.
Therefore, we first evaluate the performance of RL with different action representations on a variety of well-known benchmark tasks.
Subsequently, we apply two analysis techniques from the literature to investigate the causes of performance differences resulting from the choice of action space.
Finally, we discuss the benefits and shortcomings of each analysis method in this context.

\section{Investigating the learning performance for different action representations}
\label{sec:method}

Our analysis focuses on the well-known RL benchmark tasks from OpenAI Gym~\cite{brockman2016openai} and the DeepMind Control Suite~\cite{tunyasuvunakool2020}.
These tasks are commonly used for evaluating learning performance in RL, which allows putting our results into context.
We choose Proximal Policy Optimization~(PPO)~\cite{schulman2017proximal} as our backbone RL algorithm since it is conceptually simple and yet one of the most popular RL algorithms.

To avoid the computational expense of tuning hyperparameters for all combinations of action space and task, we use the default hyperparameters from Stable-Baselines3~\cite{raffin2021stable}.
While tuned hyperparameters are available for some tasks in~\cite{raffin2020rl}, these were optimized for the original torque-controlled versions of the tasks.
Since changing the action space might also influence which hyperparameters work well for learning, using the tuned hyperparameters would impede a fair comparison.

\subsection{Action spaces}
\label{sec:action_spaces}
We compare and analyze three action spaces: \emph{torque control}, \emph{velocity control}, and \emph{position control}.
These action spaces are widely used in robotics applications and were recently analyzed in the context of RL~\cite{peng2017learning}.
Torque control is the default actuation method for the benchmark tasks and requires no further modifications.
We implement the velocity- and position-controlled task variants by adding an additional linear controller on top of the original simulated agent.
This linear controller takes target joint positions / velocities from the RL agent and computes torques to progress toward this target.
Although it is common practice in robotics to operate such low-level controllers at a higher frequency, we execute the RL agent and the low-level controller at the same frequency to ensure a fair comparison with the torque control configuration.   %
In the following, $\pos$ and $\vel$ denote the current joint positions and velocities, respectively.
The velocity controller interprets the actions $\act{}$ of the policy as joint target velocities and computes torques according to the following control law
\begin{equation}
    \tau = K_d^{VC} (\act{} - \vel), \label{eq:velocity_control}
\end{equation}
where the controller gain $K_d^{VC}$ is a hyperparameter.

The following proportional-derivative (PD) controller implements the position control
\begin{equation}
    \tau = K_p^{PC} (\act{} - \pos) - K_d^{PC} \vel, \label{eq:position_control}
\end{equation}
where the actions $\act{}$ are interpreted as joint target positions, and $K_p^{PC}$ and $K_d^{PC}$ are controller gains that determine the influence of the position and velocity terms, respectively.
Similar to \textcite{peng2017learning}, we set the target velocity of the PD controller to zero to make the controller stabilize the system if it reaches the target position.

\begin{figure*}[htb]
    \centering
    \includegraphics[width=0.6\textwidth]{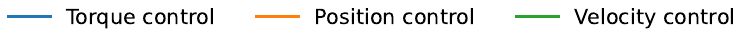} \\
    \vspace{0.05cm}
    \begin{subfigure}[b]{\dimexpr \plotfigwidth+\plotfigsep}
        \centering
        \includegraphics[width=\plotfigwidth]{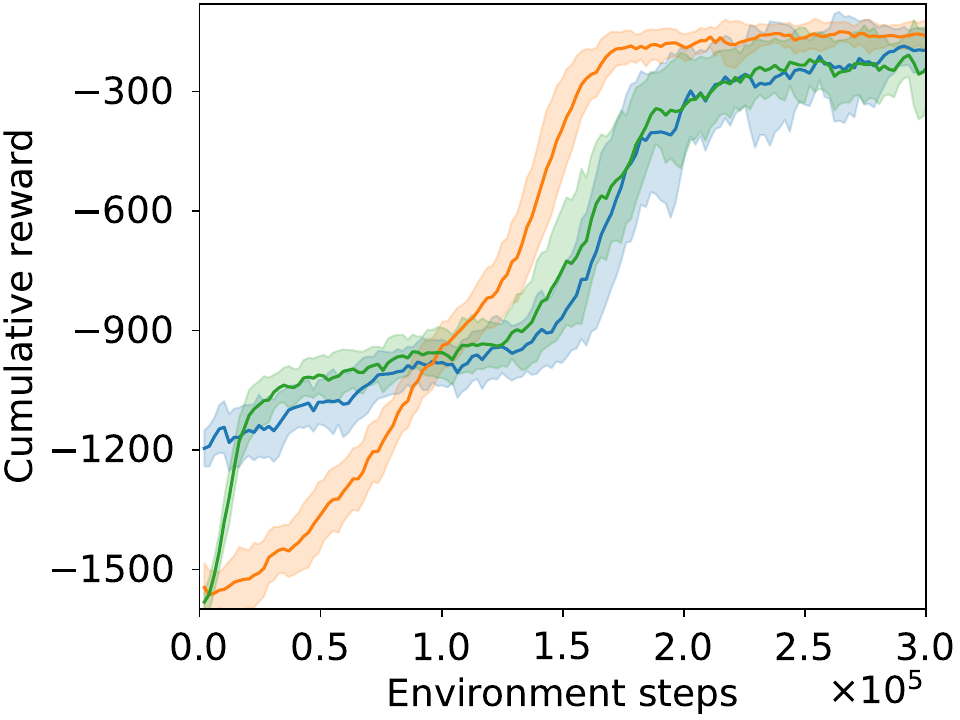}
        \caption{Gym: Pendulum}
        \label{subfig:learning_curve_gym_pendulum}
    \end{subfigure}
    \begin{subfigure}[b]{\dimexpr \plotfigwidth+\plotfigsep}
        \centering
        \includegraphics[width=\plotfigwidth]{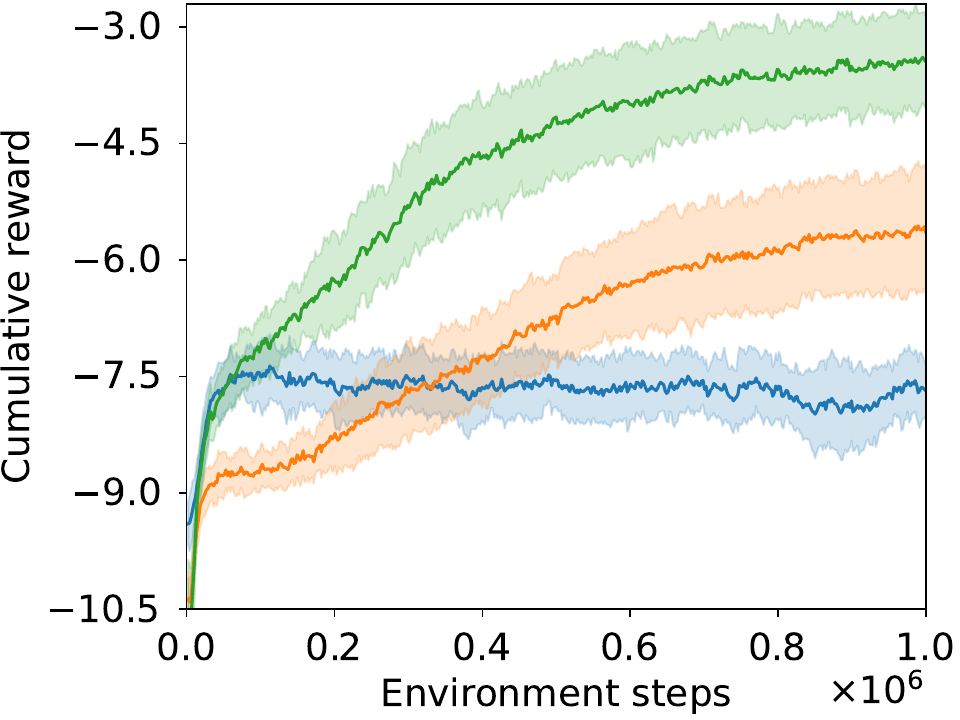}
        \caption{Gym: Reacher}
        \label{subfig:learning_curve_gym_reacher}
    \end{subfigure}
    \begin{subfigure}[b]{\dimexpr \plotfigwidth+\plotfigsep}
        \centering
        \includegraphics[width=\plotfigwidth]{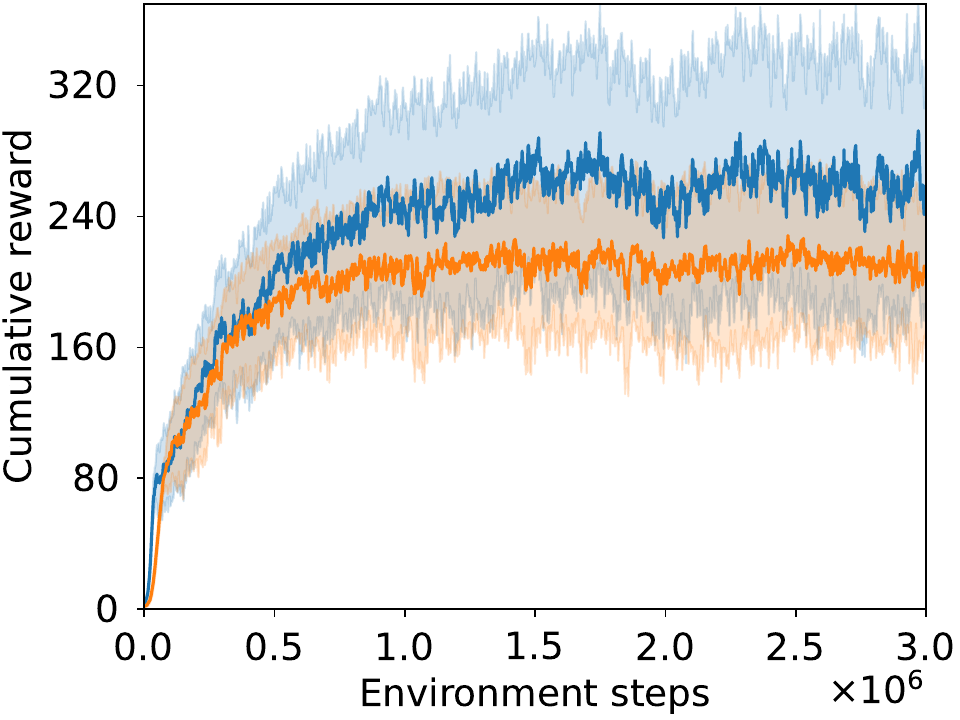}
        \caption{DMCS: Cheetah-run}
        \label{subfig:learning_curve_dmc_cheetah_run}
    \end{subfigure} \\
    \begin{subfigure}[b]{\dimexpr \plotfigwidth+\plotfigsep}
        \centering
        \includegraphics[width=\plotfigwidth]{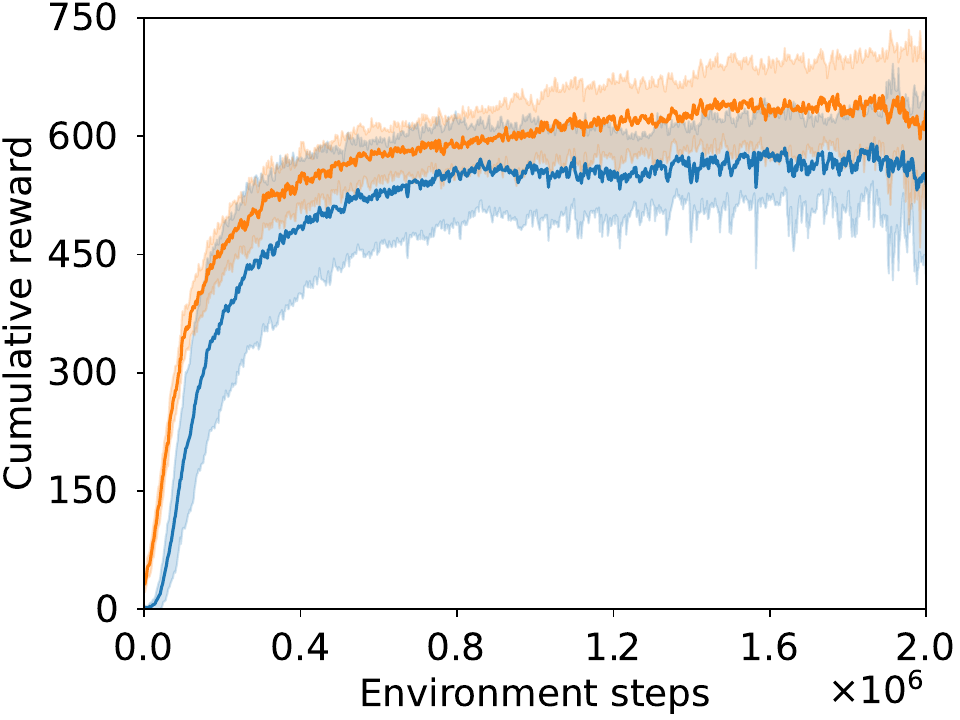}
        \caption{DMCS: Finger-spin}
        \label{subfig:learning_curve_finger_spin}
    \end{subfigure}
    \begin{subfigure}[b]{\dimexpr \plotfigwidth+\plotfigsep}
        \centering
        \includegraphics[width=\plotfigwidth]{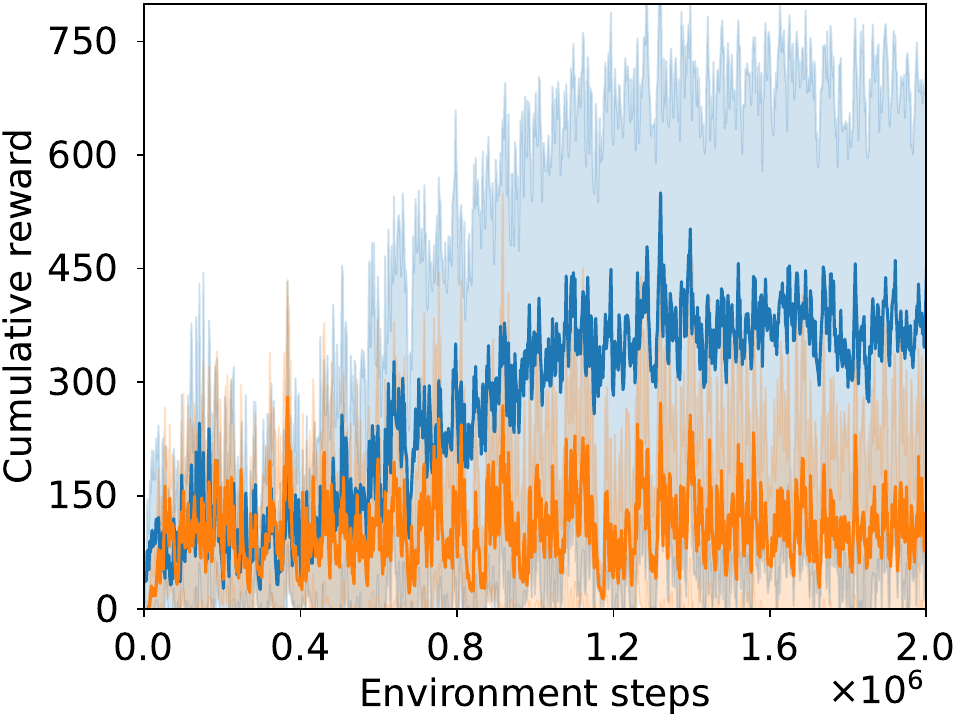}
        \caption{DMCS: Reacher-easy}
        \label{subfig:learning_curve_dmc_reacher_easy}
    \end{subfigure}
    \begin{subfigure}[b]{\dimexpr \plotfigwidth+\plotfigsep}
        \centering
        \includegraphics[width=\plotfigwidth]{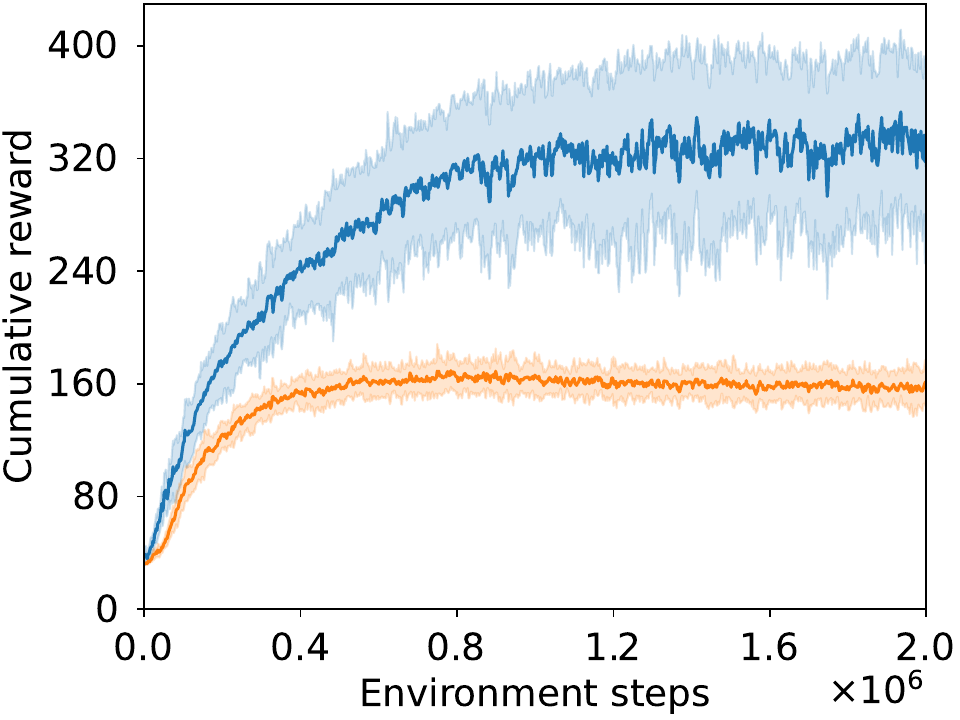}
        \caption{DMCS: Walker-walk}
        \label{subfig:learning_curve_dmc_walker_walk}
    \end{subfigure}
    \caption{
        Learning curves of PPO on benchmark tasks from OpenAI Gym~\cite{brockman2016openai} and the DeepMind Control Suite~(DMCS)~\cite{tunyasuvunakool2020} with different action spaces.
        All results are averaged over 10 random seeds with the shaded area representing the standard deviation over the seeds.
        For the tasks Pendulum, Reacher, and Walker-walk, the choice of the action space has a significant impact on the learning performance.
        The performance difference is smaller but still noticeable for the remaining tasks.
    }
    \label{fig:learning_curves}
\end{figure*}

When using these action representations for RL, we observe significant differences in learning performance.
The learning curves depicted in \cref{fig:learning_curves} illustrate this observation.
However, there is no action space that works best across all tasks.
To understand the causes for the performance differences, we take a closer look at different aspects of the policy optimization and show analysis results for two tasks: Reacher from OpenAI Gym and Walker-walk from the DeepMind Control Suite.
The tasks require distinct skills and the learning performance is significantly affected by the choice of action space.
We refer to \cref{app:results_other_tasks} for results on the remaining tasks.

\subsection{Visualizing the optimization landscape does not always yield insights into learning performance}
\label{subsec:optimization_surface_visualization}

Generally, the shape of the optimization surface determines the difficulty of a learning problem.
A rugged optimization landscape with many shallow local optima is usually more challenging to navigate than a smooth, convex one.
Gradient-based methods, such as policy gradients, are particularly affected by the shape of the optimization surface since they are prone to converge to sub-optimal solutions.

Visual inspection of loss landscapes can provide intuitive insights into the difficulty of an optimization problem. 
However, due to the large number of parameters in typical neural networks, it is impossible to visualize the loss directly.
\textcite{li2018visualizing} propose a dimensionality reduction technique to create two-dimensional visualizations of the loss surface of neural networks.
Starting from the current parameters, the technique plots the value of the loss function along two random directions in parameter space.
Recently, \textcite{sullivan2022cliff,bekci2020visualizing} applied the method to analyze RL algorithms.

Similar to \textcite{sullivan2022cliff}, we discretize the parameter space on a $31 \times 31$ grid along the two random directions.
For each point on the grid, we create an instance of the agent with parameters corresponding to the grid location.
We evaluate the discounted cumulative reward and the PPO loss of each agent with $200{,}000$ samples to obtain good approximations of the true values of these objectives.
Even though the visualization method includes a significant amount of stochasticity, we find that plots generated with different random seeds are visually similar, see \cref{app:opt_vis_reproducibility}.

\begin{figure*}[htb]
    \centering
    \begin{subfigure}{0.6\textwidth}
        \centering
        \begin{subsubfigures}
        \begin{subsubfigure}{\surfacefigwidth}
            \includegraphics[width=\textwidth]{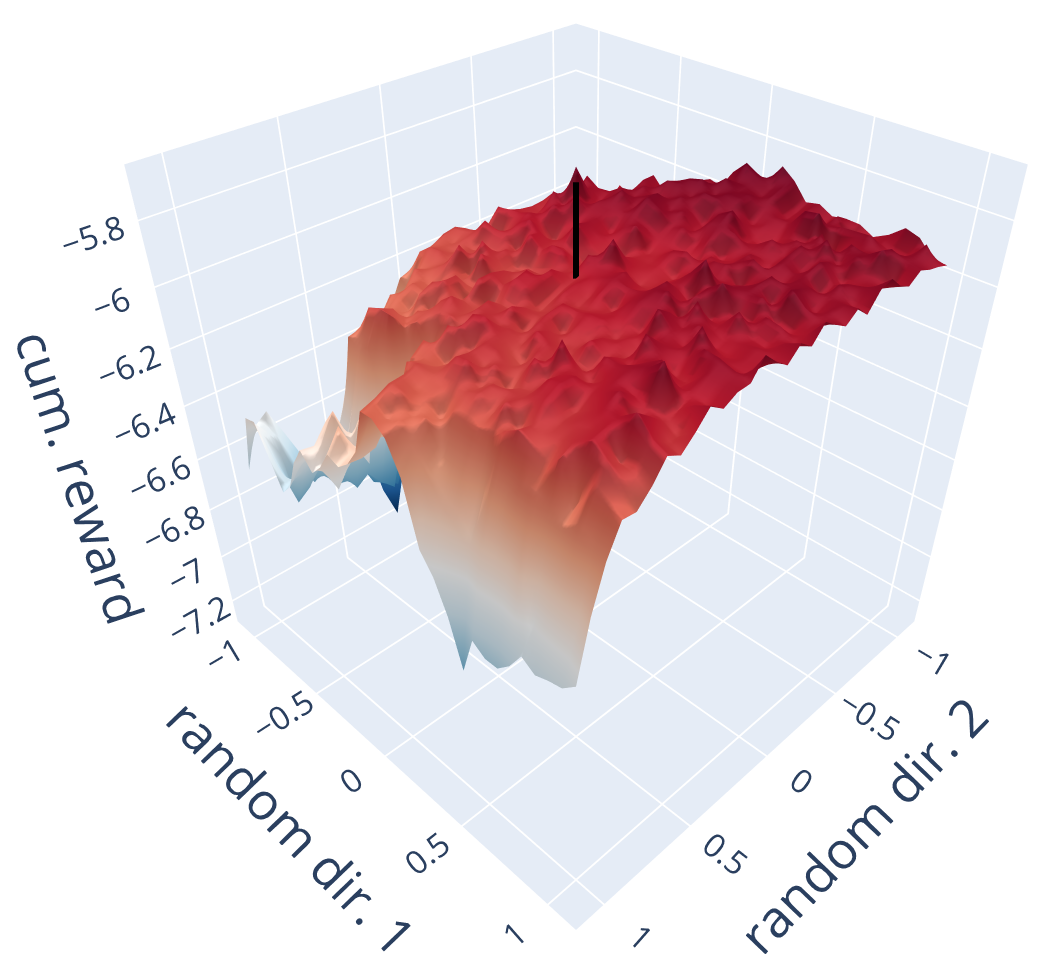}
            \caption{TC, reward}
            \label{subfig:opt_vis_gym_reacher_tc_reward}
        \end{subsubfigure}
        \begin{subsubfigure}{\surfacefigwidth}
            \includegraphics[width=\textwidth]{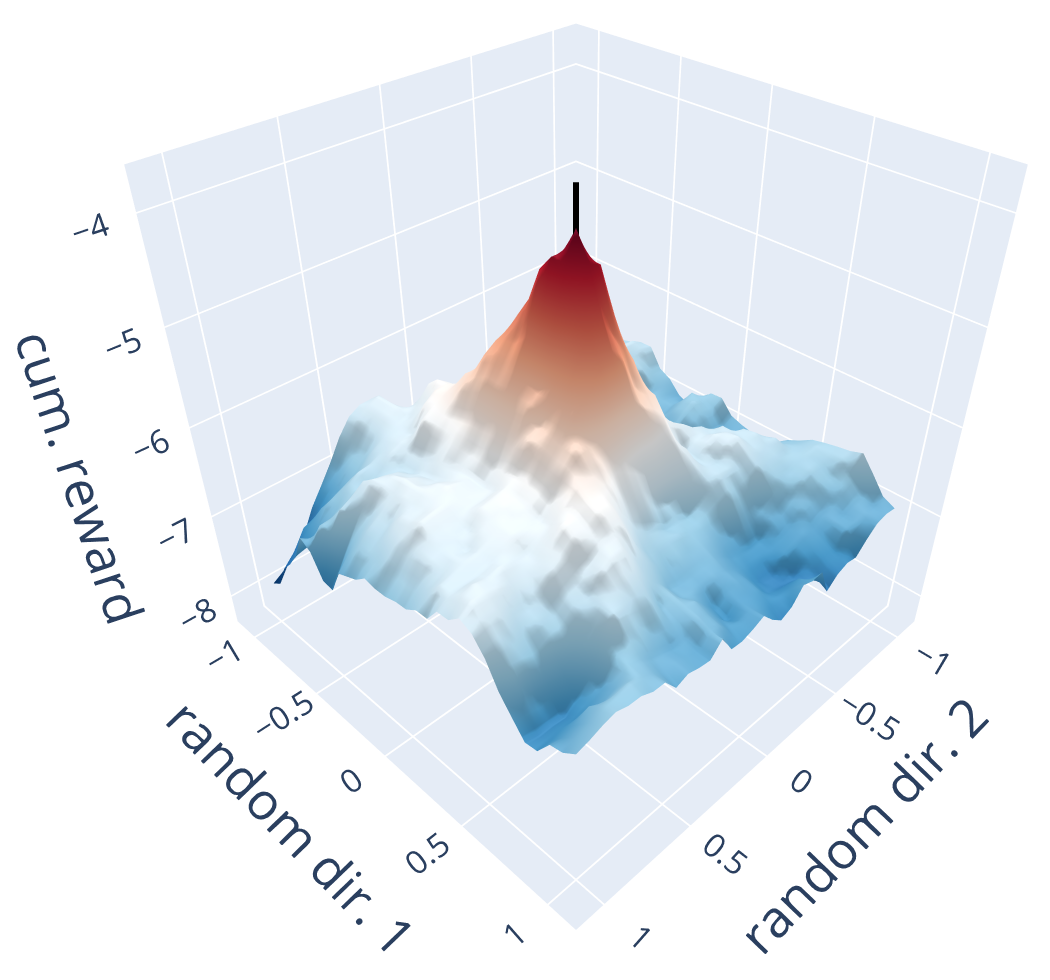}
            \caption{PC, reward}
            \label{subfig:opt_vis_gym_reacher_pc_reward}
        \end{subsubfigure}
        \begin{subsubfigure}{\surfacefigwidth}
            \includegraphics[width=\textwidth]{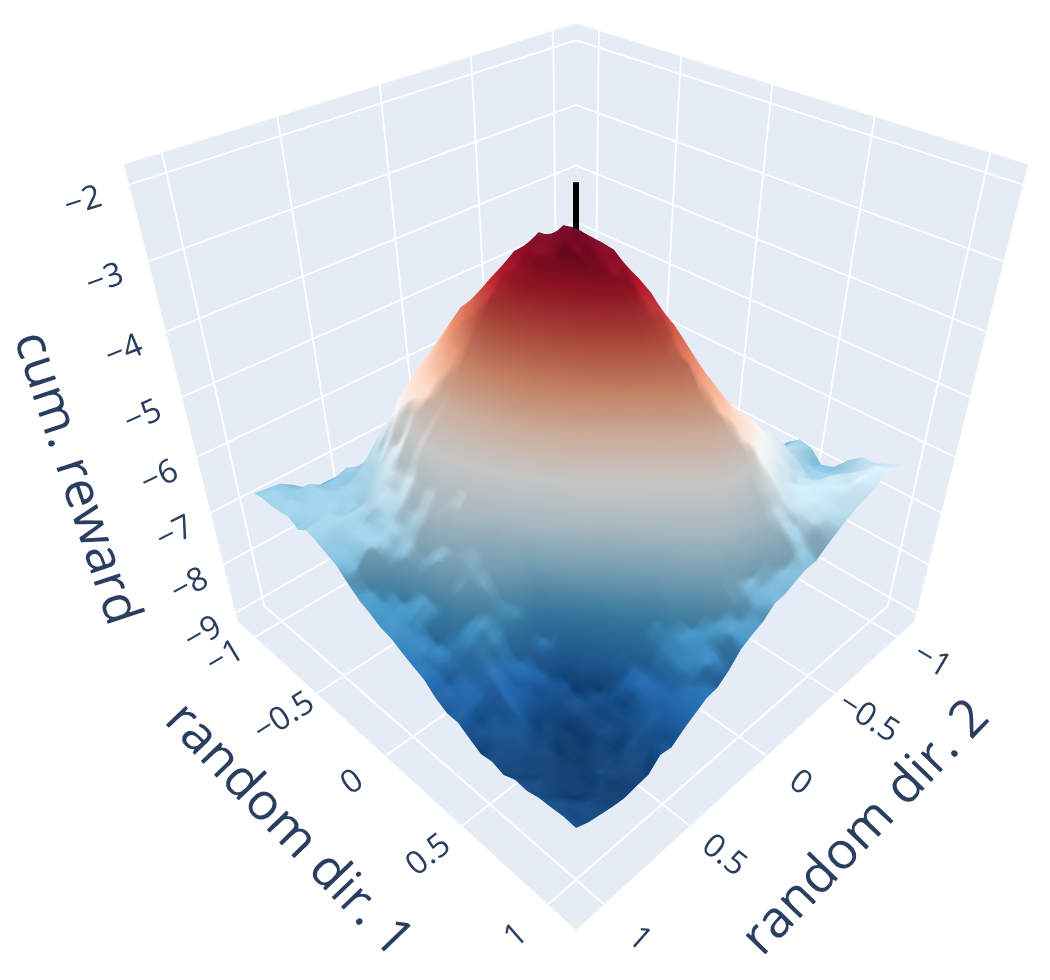}
            \caption{VC, reward}
            \label{subfig:opt_vis_gym_reacher_vc_reward}
        \end{subsubfigure} \\
        \begin{subsubfigure}{\surfacefigwidth}
            \includegraphics[width=\textwidth]{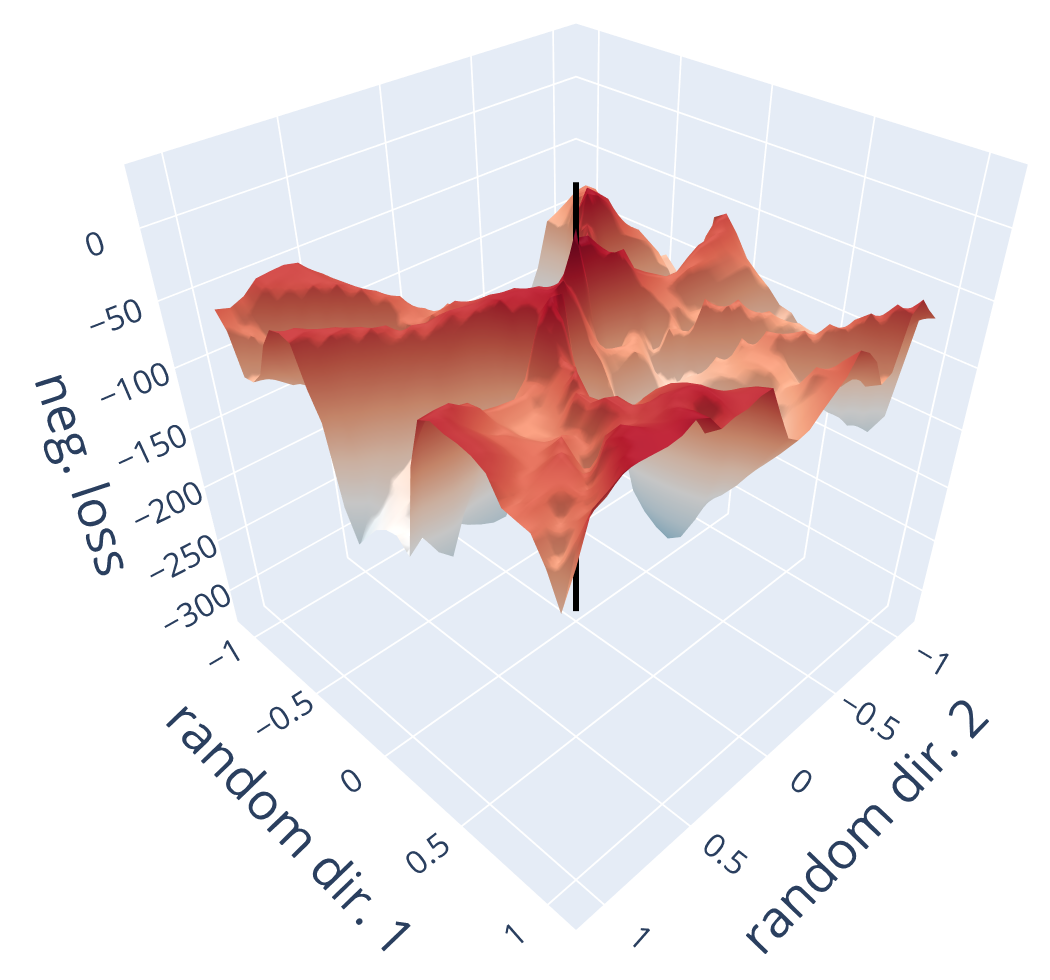}
            \caption{TC, neg.\ loss}
            \label{subfig:opt_vis_gym_reacher_tc_combined}
        \end{subsubfigure}
        \begin{subsubfigure}{\surfacefigwidth}
            \includegraphics[width=\textwidth]{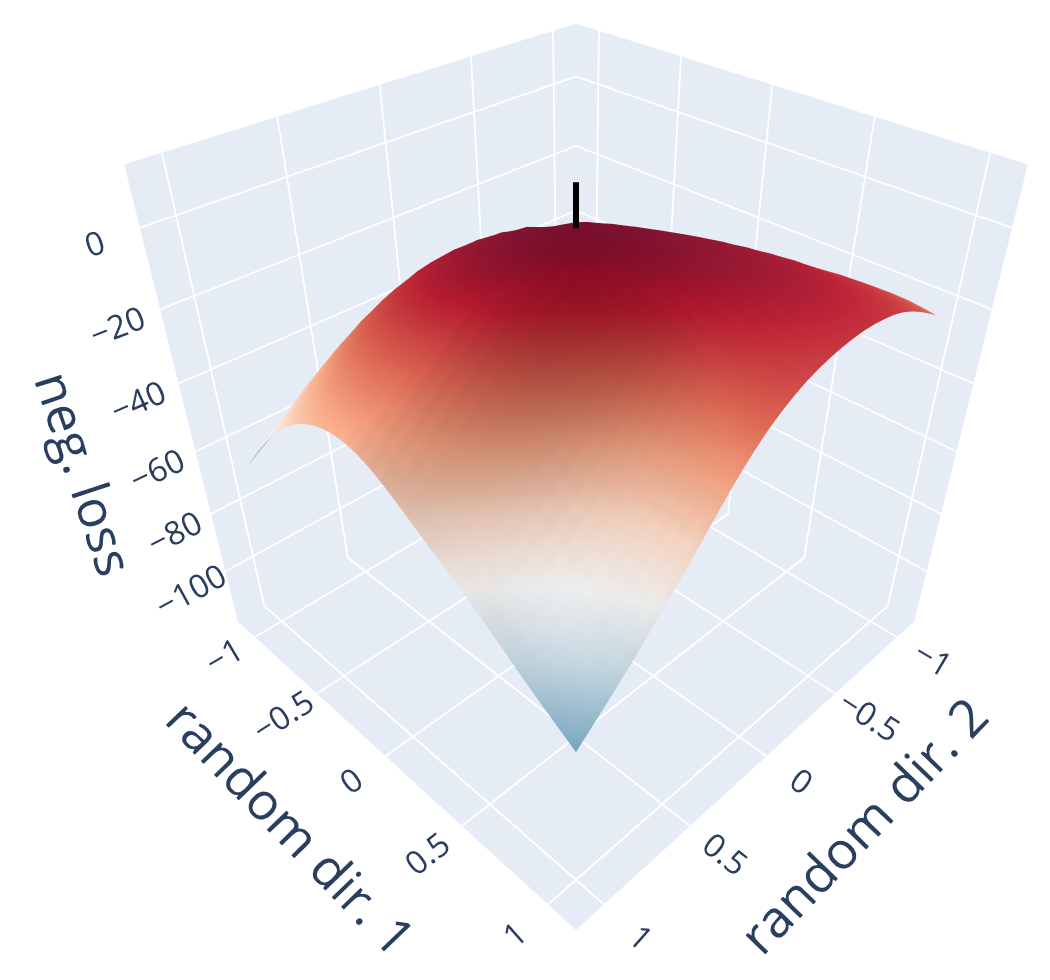}
            \caption{PC, neg.\ loss}
            \label{subfig:opt_vis_gym_reacher_pc_combined}
        \end{subsubfigure}
        \begin{subsubfigure}{\surfacefigwidth}
            \includegraphics[width=\textwidth]{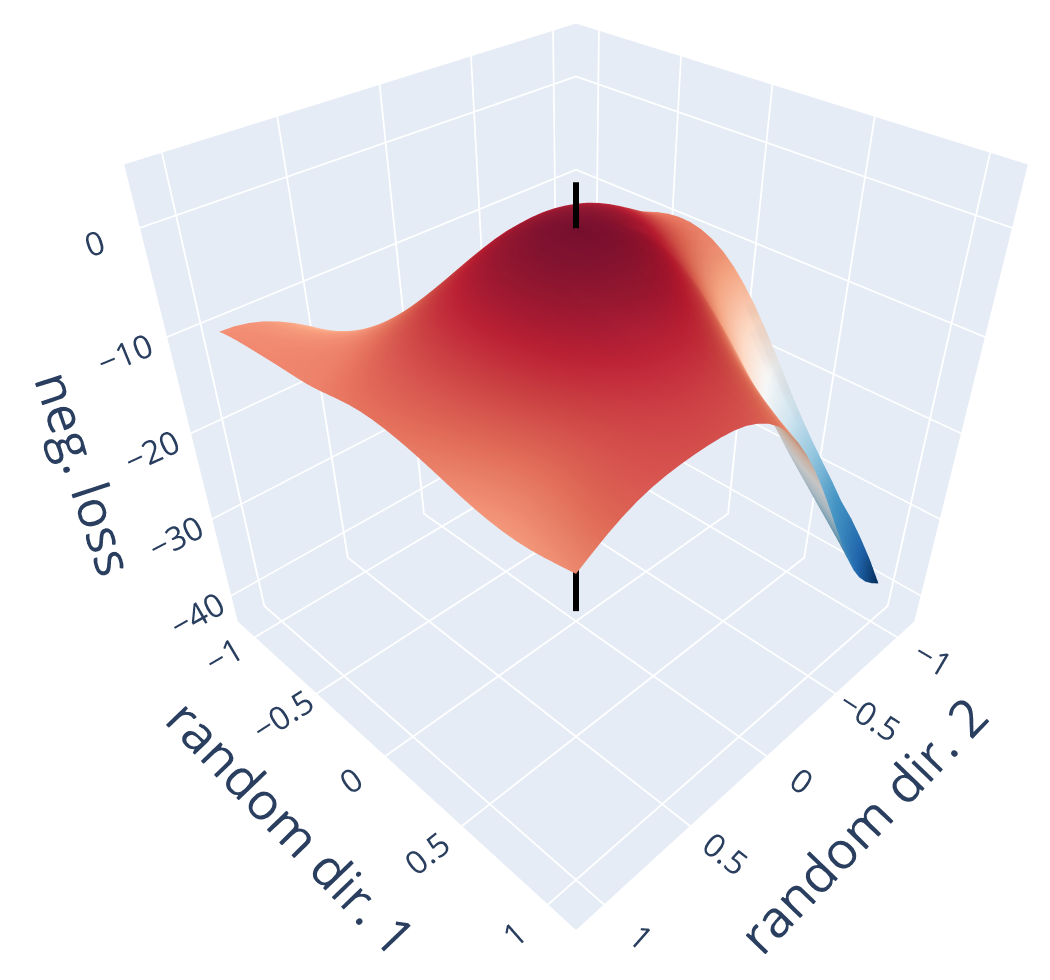}
            \caption{VC, neg.\ loss}
            \label{subfig:opt_vis_gym_reacher_vc_combined}
        \end{subsubfigure}
        \end{subsubfigures}
        \caption{Visualizations for Reacher}
        \label{subfig:opt_vis_gym_reacher}
    \end{subfigure}
    \hfill
    \begin{subfigure}{0.39\textwidth}
        \centering
        \begin{subsubfigures}
            \begin{subsubfigure}{\surfacefigwidth}
                \includegraphics[width=\textwidth]{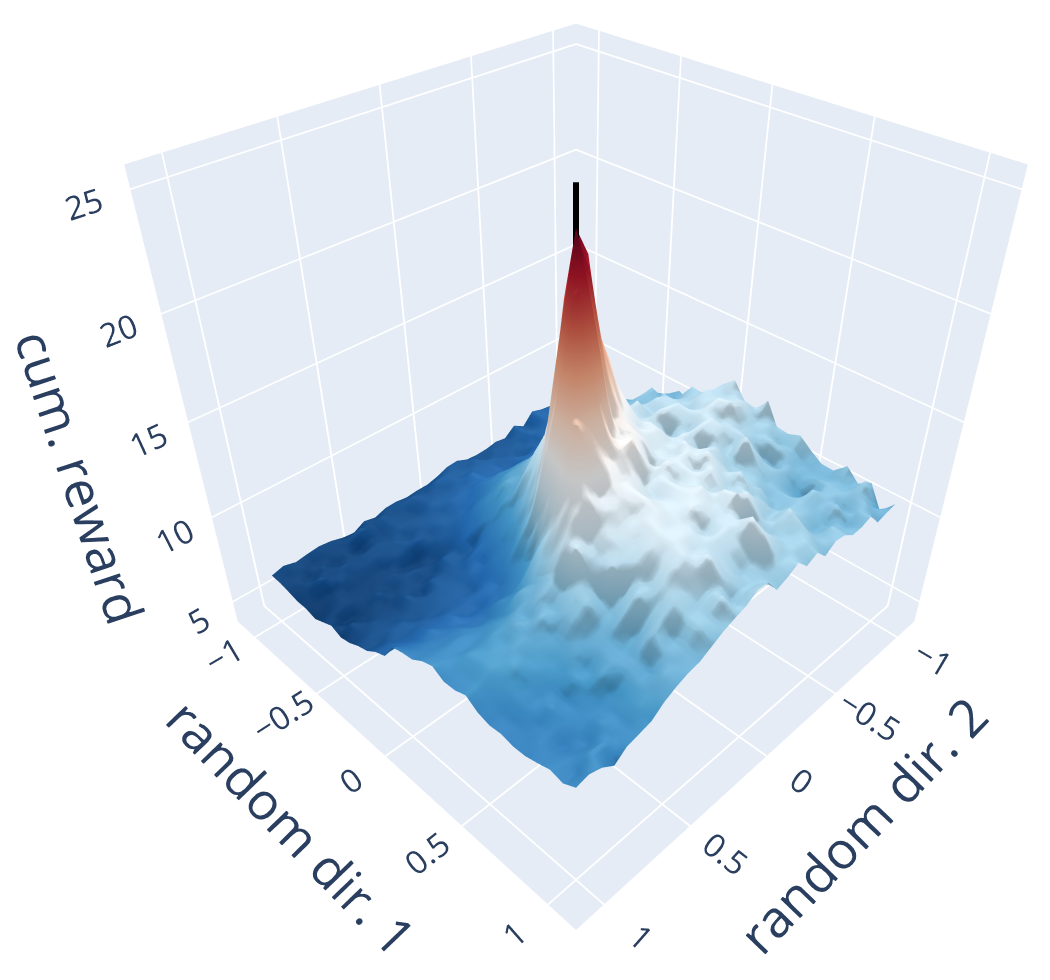}
                \caption{TC, reward}
                \label{subfig:opt_vis_dmc_walker_walk_tc_reward}
            \end{subsubfigure}
            \begin{subsubfigure}{\surfacefigwidth}
                \includegraphics[width=\textwidth]{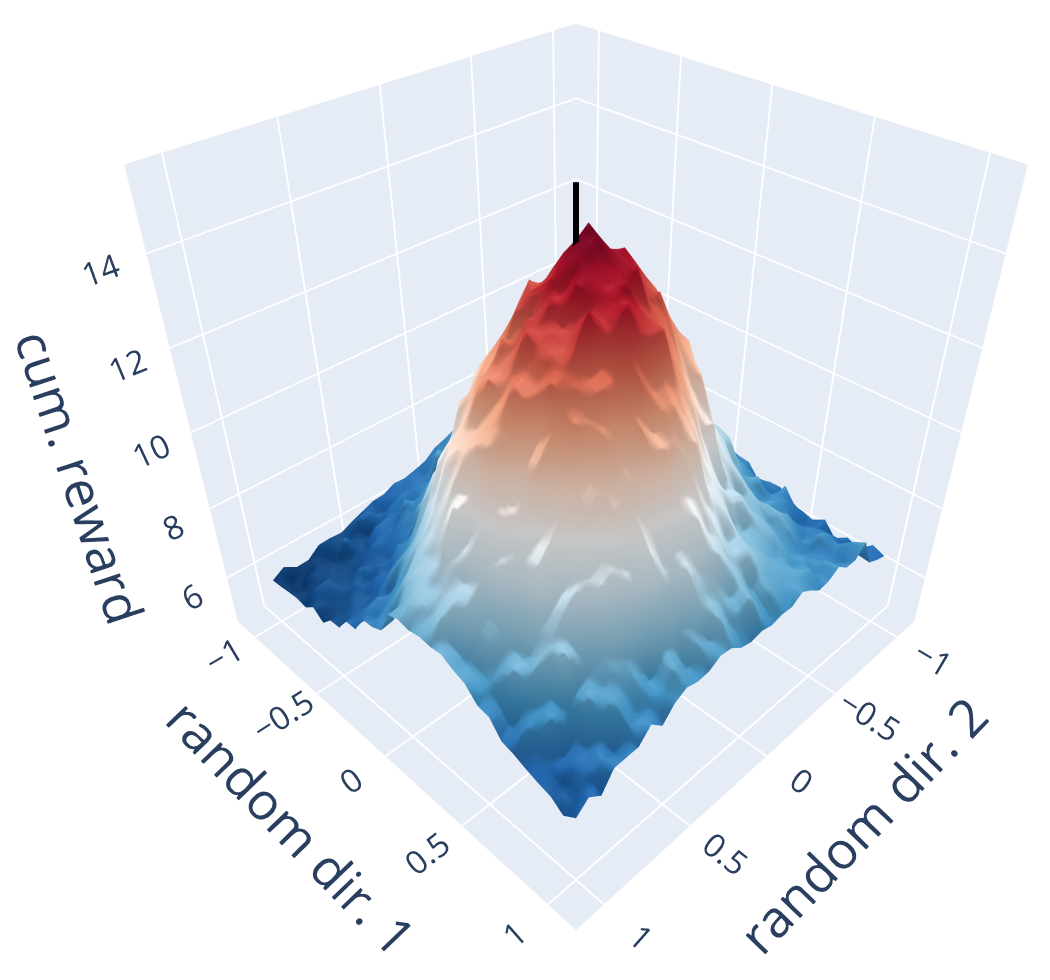}
                \caption{PC, reward}
                \label{subfig:opt_vis_dmc_walker_walk_pc_reward}
            \end{subsubfigure} \\
            \begin{subsubfigure}{\surfacefigwidth}
                \includegraphics[width=\textwidth]{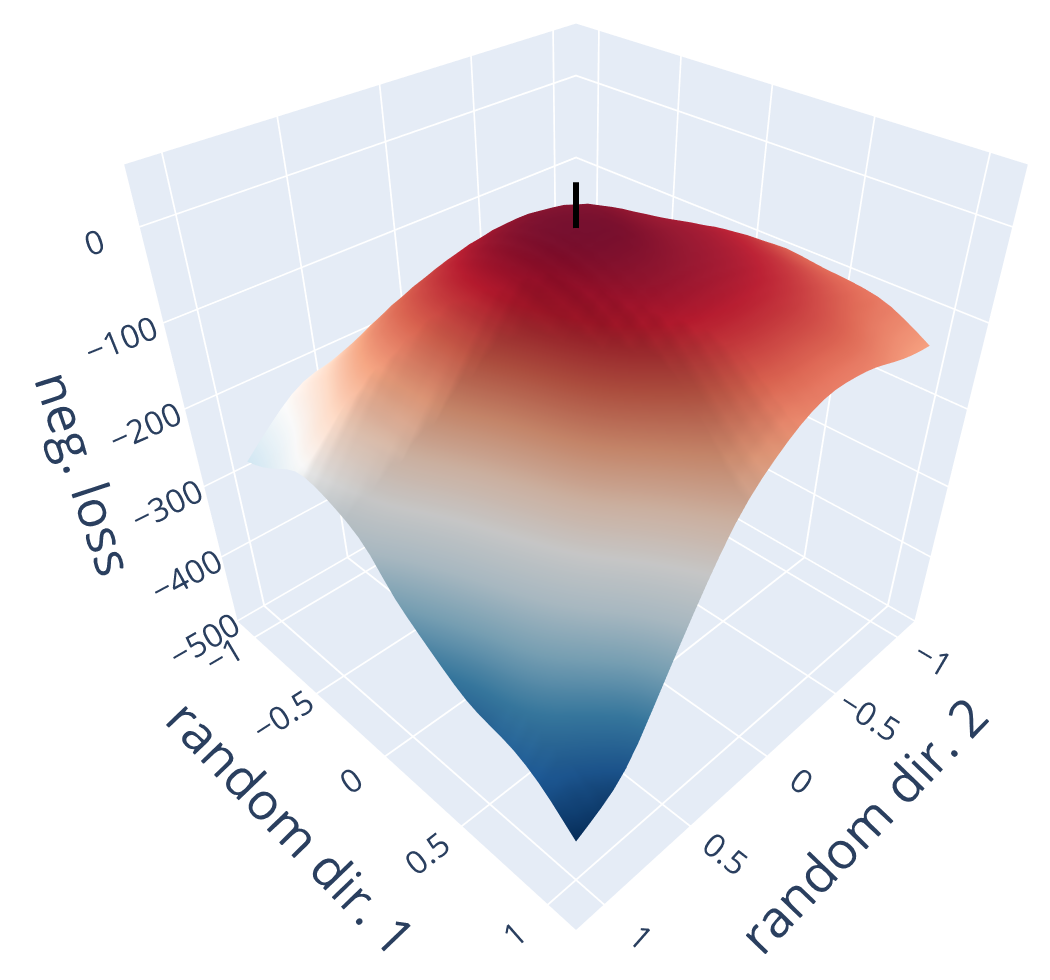}
                \caption{TC, neg.\ loss}
                \label{subfig:opt_vis_dmc_walker_walk_tc_combined}
            \end{subsubfigure}
            \begin{subsubfigure}{\surfacefigwidth}
                \includegraphics[width=\textwidth]{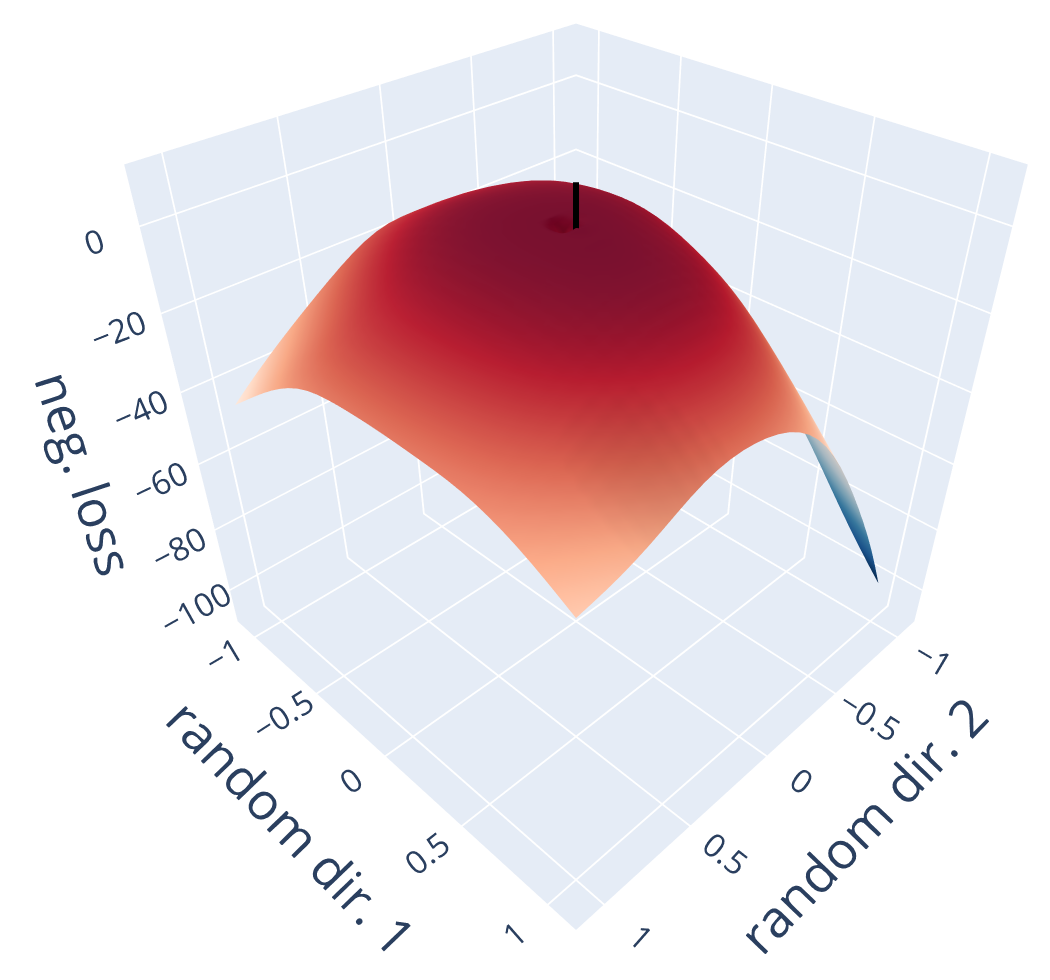}
                \caption{PC, neg.\ loss}
                \label{subfig:opt_vis_dmc_walker_walk_pc_combined}
            \end{subsubfigure}
        \end{subsubfigures}
        \caption{Visualizations for Walker-walk}
        \label{subfig:opt_vis_dmc_walker_walk}
    \end{subfigure}
    \caption{
        Visualizations of the optimization surface for the Reacher (left) and Walker-walk (right) tasks.
        All plots are generated by plotting the objective value along two random directions in parameter space, starting from the parameters found after training for 1 million environment steps (marked by the black line).
        The top row shows the discounted cumulative reward, while the bottom row displays the surrogate loss optimized by PPO (negated for visual clarity).
        For Reacher, the optimization surface of the torque control~(TC) configuration (plots \subref{subfig:opt_vis_gym_reacher_tc_reward} and \subref{subfig:opt_vis_gym_reacher_tc_combined}) appears significantly more complex than that of the position control~(PC) and velocity control~(VC) configurations (plots \subref{subfig:opt_vis_gym_reacher_pc_reward}, \subref{subfig:opt_vis_gym_reacher_pc_combined} and plots \subref{subfig:opt_vis_gym_reacher_vc_reward}, \subref{subfig:opt_vis_gym_reacher_vc_combined}, respectively).
        For Walker-walk, the complexity of the optimization surfaces for the two action representations appears similar.
    }
    \label{fig:opt_vis}
\end{figure*}

PPO does not directly optimize the cumulative reward and operates on a surrogate loss instead.
Some works analyze the reward surface~\cite{sullivan2022cliff}, while others focus on the loss surface~\cite{ilyas2020closer,bekci2020visualizing}.
To get the complete picture, we visualize both the reward and loss surfaces resulting from different action spaces in \cref{fig:opt_vis}.
\Cref{subfig:opt_vis_gym_reacher_tc_combined} shows that the loss surface for the Reacher torque control configuration is rugged and has numerous local optima.
The difficulty of optimizing such a loss might explain the poor learning performance of the agent in this configuration.
For the remaining action representations, the optimization surface appears simpler to optimize.
\Cref{subfig:opt_vis_dmc_walker_walk} does not yield a clear intuition for the performance difference between the two Walker-walk configurations.
The loss surfaces are visually very similar, yet the agent performs significantly better in the torque control configuration.

Another interesting observation is the striking discrepancy between the visualizations of the reward surfaces and the loss surfaces.
This observation supports the finding in~\cite{ilyas2020closer} that the loss that PPO optimizes is often poorly correlated with the cumulative reward, which is the true objective in RL.

We showed that optimization surface visualizations can yield intuitive insights into the optimization difficulty for certain tasks.
However, in other cases, the interpretation of the results is not as clear since the method does not produce a clear metric for comparison.
Furthermore, this approach is constrained to analyzing a single snapshot of the training at a time.
Due to the continual update of the policy, the distribution of collected data changes constantly, which affects the loss landscape. %
Moreover, the method uses a large number of samples to approximate the true objective.
However, during RL training, the optimizer does not directly operate on the true objective but rather uses a coarse approximation generated with a low number of samples.
Given these limitations, we describe an alternative analysis method in the following section.
This method results in a numerical criterion, which facilitates comparison across different configurations and allows visualizing results for the entire training as a simple line plot.

\subsection{More accurate gradient estimates do not imply better learning performance}
\label{subsec:gradient_analysis}

In this section, we explore the effects that the low-sample approximation of the loss has on the gradient estimation.
The accuracy of the gradient estimates is crucial since it determines the direction of the policy update.
Precise gradient estimates allow reaching an optimum of the objective quickly, while noisy estimates can result in the optimization oscillating through the loss landscape without improving the agent's performance.
We utilize the analysis method by \textcite{ilyas2020closer} to measure and compare the quality of the gradient estimates in tasks with different action spaces.
To evaluate the gradient quality throughout the entire learning process, we execute the analysis at 20 checkpoints during the agent's training.
For each checkpoint, we approximate the true gradient with $10^7$ samples.
In contrast, PPO with default hyperparameters uses $64$ samples to estimate the gradient during training.
We compute the average pairwise cosine similarity between the gradients estimated during training and our approximation of the true gradient.
This quantity provides insights into how closely the gradients used during training match the true gradient.
Since the PPO loss is the sum of a policy loss and a value function loss, we also execute the analysis on each term individually to measure which of the terms is more problematic for the gradient estimation.

\begin{figure*}[htb]
    \centering
    \includegraphics[width=0.6\textwidth]{figures/legend.pdf} \\
    \vspace{0.05cm}
    \begin{subfigure}[b]{\dimexpr \plotfigwidth+\plotfigsep}
        \centering
        \includegraphics[width=\plotfigwidth]{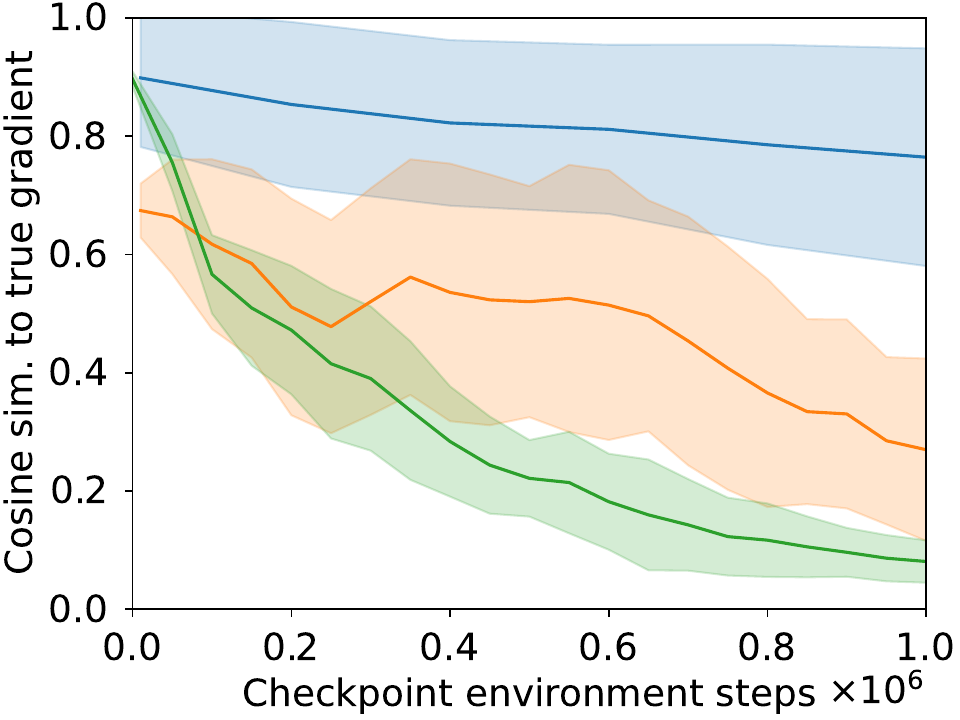}
        \caption{Reacher, total loss}
        \label{subfig:gradient_sim_true_gym_reacher_combined}
    \end{subfigure}
    \begin{subfigure}[b]{\dimexpr \plotfigwidth+\plotfigsep}
        \centering
        \includegraphics[width=\plotfigwidth]{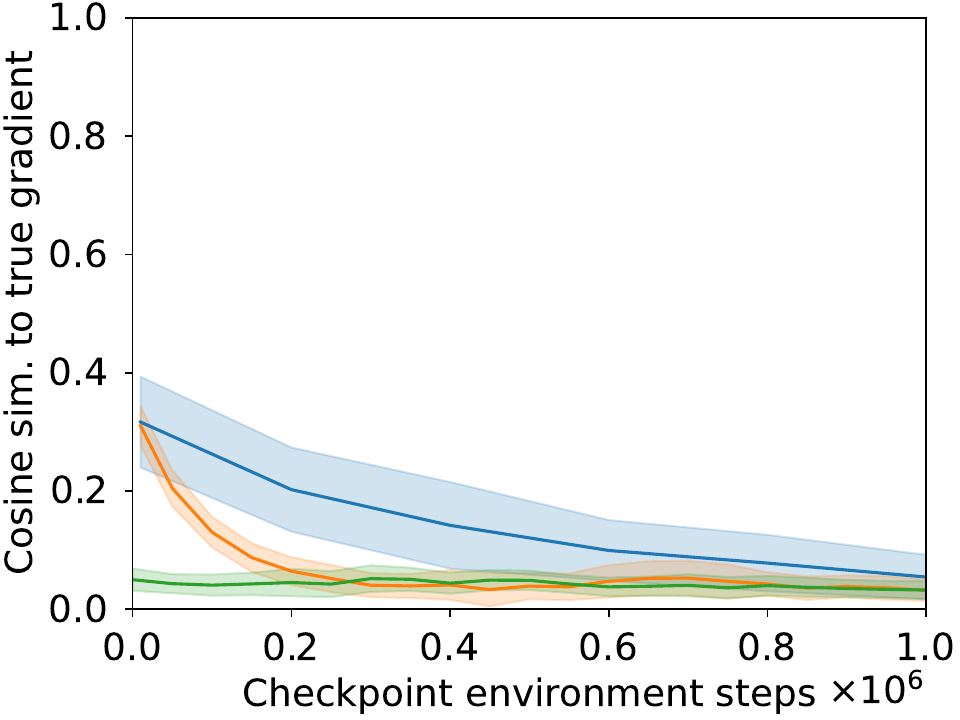}
        \caption{Reacher, policy loss}
        \label{subfig:gradient_sim_true_gym_reacher_policy}
    \end{subfigure}
    \begin{subfigure}[b]{\dimexpr \plotfigwidth+\plotfigsep}
        \centering
        \includegraphics[width=\plotfigwidth]{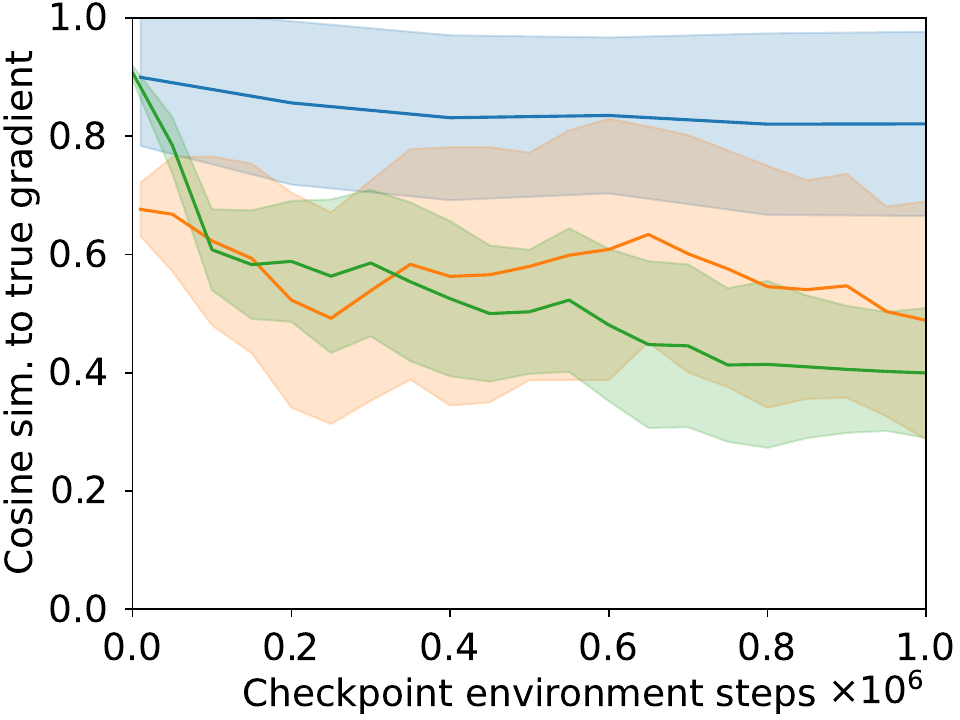}
        \caption{Reacher, value function loss}
        \label{subfig:gradient_sim_true_gym_reacher_vf}
    \end{subfigure} \\
    \begin{subfigure}[b]{\dimexpr \plotfigwidth+\plotfigsep}
        \centering
        \includegraphics[width=\plotfigwidth]{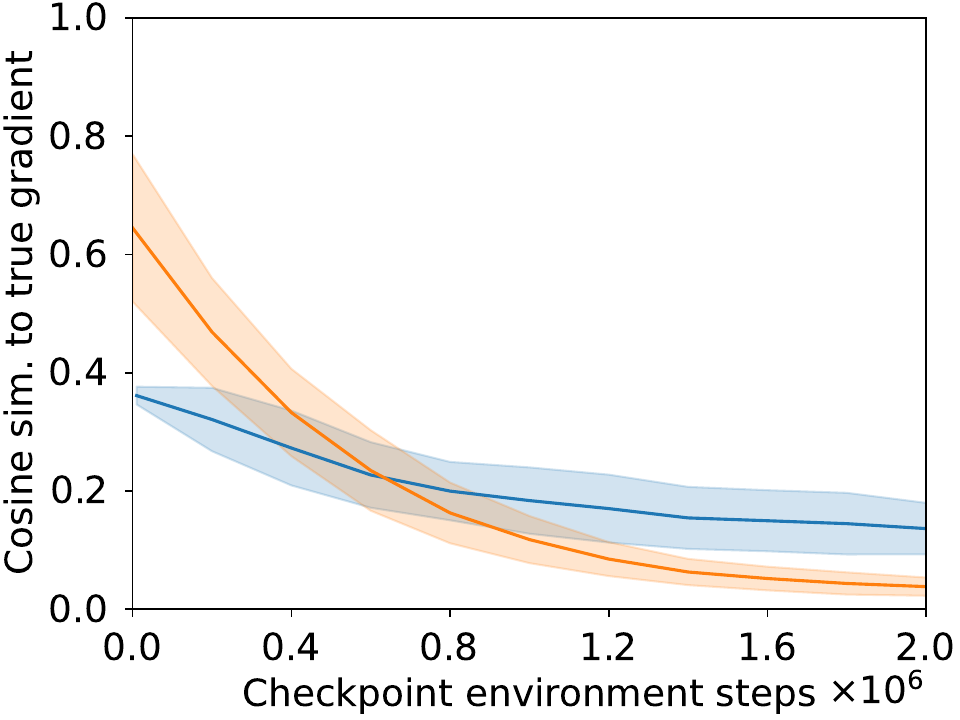}
        \caption{Walker-walk, total loss}
        \label{subfig:gradient_sim_true_dmc_walker_walk_combined}
    \end{subfigure}
    \begin{subfigure}[b]{\dimexpr \plotfigwidth+\plotfigsep}
        \centering
        \includegraphics[width=\plotfigwidth]{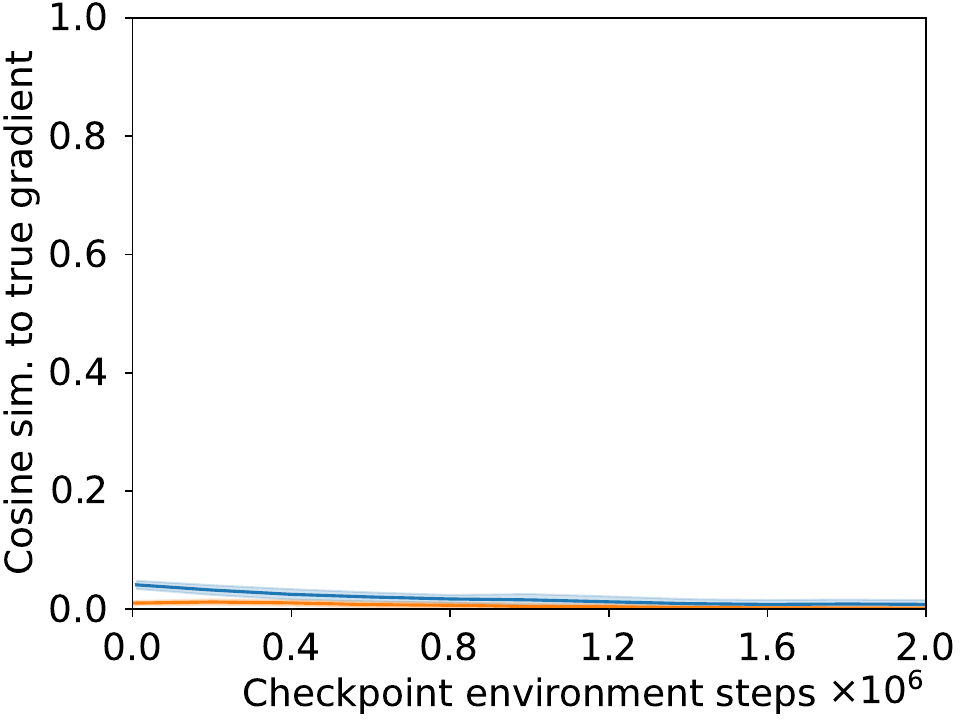}
        \caption{Walker-walk, policy loss}
        \label{subfig:gradient_sim_true_dmc_walker_walk_policy}
    \end{subfigure}
    \begin{subfigure}[b]{\dimexpr \plotfigwidth+\plotfigsep}
        \centering
        \includegraphics[width=\plotfigwidth]{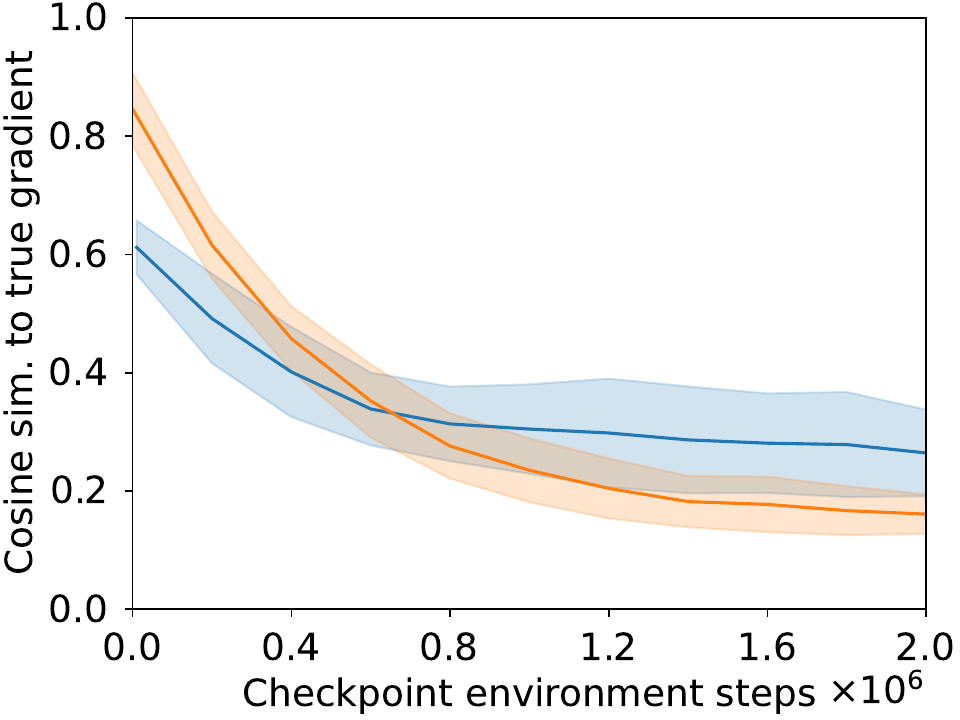}
        \caption{Walker-walk, value function loss}
        \label{subfig:gradient_sim_true_dmc_walker_walk_vf}
    \end{subfigure}
    \caption{
        The quality of the gradient estimates measured as the average cosine similarity to a good approximation of the true gradient, computed with $10^7$ samples.
        All plots show the analysis results at 20 checkpoints during training, averaged over 10 training runs with different random seeds.
        The top row displays the results on the Reacher task, while the bottom row focuses on Walker-walk.
        Plots \subref{subfig:gradient_sim_true_gym_reacher_combined} and \subref{subfig:gradient_sim_true_dmc_walker_walk_combined} display the results for the gradients estimated for the entire PPO loss and plots \subref{subfig:gradient_sim_true_gym_reacher_policy}, \subref{subfig:gradient_sim_true_dmc_walker_walk_policy} and \subref{subfig:gradient_sim_true_gym_reacher_vf}, \subref{subfig:gradient_sim_true_dmc_walker_walk_vf} respectively regard the policy and value function loss individually.
        For Reacher, the better-performing configurations use worse gradient estimates.
        For Walker-walk, this trend reverses at around $700{,}000$ steps.
        Furthermore, the gradient quality is significantly worse for the policy loss than for the value function loss.
    }
    \label{fig:gradient_sim_true}
\end{figure*}

\Cref{fig:gradient_sim_true} displays the results of the gradient analysis for the Reacher and Walker-walk tasks.
The results appear counter-intuitive since the better-performing configurations for Reacher use worse gradients.
For Walker-walk, this is also the case for the first $700{,}000$ steps.
Afterward, the trend reverses, and the better-performing configuration also uses better gradient estimates.
We hypothesize that these unclear results are caused by a more complex relationship between the gradient quality and learning performance.
Not only does the gradient estimation influence the learning performance, but the learning performance also affects the gradient estimation.
If the optimizer gets close to an optimum, the loss landscape locally becomes relatively flat, and small estimation errors can change the gradient direction drastically.
Furthermore, noisy gradient estimates can even be beneficial for escaping shallow local optima~\cite{neelakantan2015adding}.
In \cref{app:gt_gradient_training}, we demonstrate that training with more accurate gradient estimates does not always yield performance gains.

Another interesting observation in \cref{fig:gradient_sim_true} is that the gradient accuracy is significantly worse for the policy loss than for the value function loss.
A plausible explanation is the use of importance sampling in the policy loss, which increases the influence of individual samples on the loss term.

\section{Conclusion}
\label{sec:conclusion}

In this work, we demonstrated that action representations can have a significant impact on the learning performance for RL benchmark tasks.
We described two analysis methods and evaluated their effectiveness for analyzing the impact of the action representation.
Our results suggest that the performance gains for the Reacher task can be attributed to a smoother optimization landscape but this explanation does not seem to transfer to Walker-walk, demonstrating that visualizing the optimization landscape is not always sufficient to gather insights about the learning performance.
Furthermore, our evaluation indicates that the accuracy of the gradient estimates does not correlate well with learning performance, which suggests a more complex relationship between these two factors.

Our analysis results are not yet sufficient to explain all observed performance differences.
To achieve this goal, the following challenges need to be addressed.
\numericparagraph{Normalizing the analysis results with respect to the learning progress}
Properties like the complexity of the optimization landscape and the accuracy of gradient estimates influence how quickly the agent progresses on the objective.
However, agents on different levels of performance generally operate in different parts of the state space, which in turn can have an effect on these properties.
Proficient agents might operate in regions of the state space that require fine-grained control, which in turn could result in optimization surfaces that are challenging to optimize. 
To fully understand influences on learning performance, it is crucial to control for these effects.

\numericparagraph{Disentangling different effects on the RL algorithm}
Changing the action representation typically affects multiple aspects of the learning process simultaneously.
For example, we have seen the influence on both the optimization landscape complexity and the gradient estimation accuracy in \cref{subsec:optimization_surface_visualization,subsec:gradient_analysis}, respectively.
Furthermore, such changes could also affect other algorithm components, like the agent's exploration.
Since these effects typically occur together, it remains unclear which influences are responsible for the performance differences.

\numericparagraph{Assessing the effects of other task design choices and hyperparameters}
Other choices in the task design, like the state representation, also have a significant impact on the learning performance and analysis results.
Specifically, there might be choices that are particularly suitable for certain action spaces.
In \cref{app:states_and_actions_in_the_same_space}, we investigate the effect of defining both the states and actions in the same space.
Similarly, the hyperparameters of the learning algorithm and the controller gains influence the results.
We tuned the controller gains to minimize the average error over a fixed number of time steps.
However, the reinforcement learning algorithm might yield better results with a more aggressive controller since this choice potentially allows to utilize higher torques.

Further understanding the impact of action representations on learning performance enables tuning action representations to make learning more efficient and robust.
Beyond the use for optimizing action representations, sophisticated analysis tools could also provide vital insights into obstacles that impede learning progress in practice.
These insights could pave the way for designing more sample-efficient and robust RL algorithms.%
\pagebreak %

\FloatBarrier

\enlargethispage{-7cm}
\printbibliography

\newpage

\onecolumn

\appendix

\subsection{Analysis results on remaining tasks}
\label{app:results_other_tasks}

\subsubsection{Optimization surface visualization}
\mbox{}     %

\begin{figure}[H]
    \centering
    \begin{subfigure}[b]{\surfacefigwidthappendix}
        \centering
        \includegraphics[width=\textwidth]{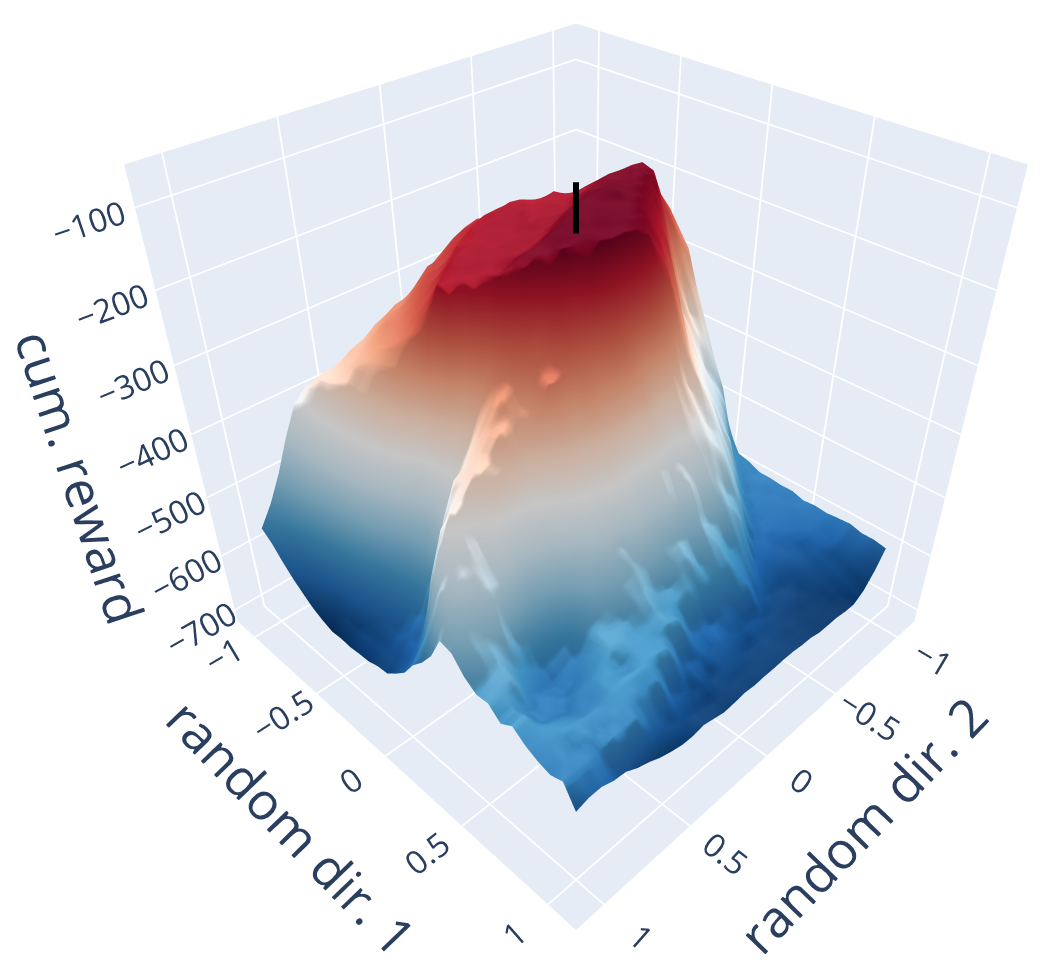}
        \caption{Torque control, reward}
        \label{subfig:opt_vis_gym_pendulum_tc_reward}
    \end{subfigure}
    \begin{subfigure}[b]{\surfacefigwidthappendix}
        \centering
        \includegraphics[width=\textwidth]{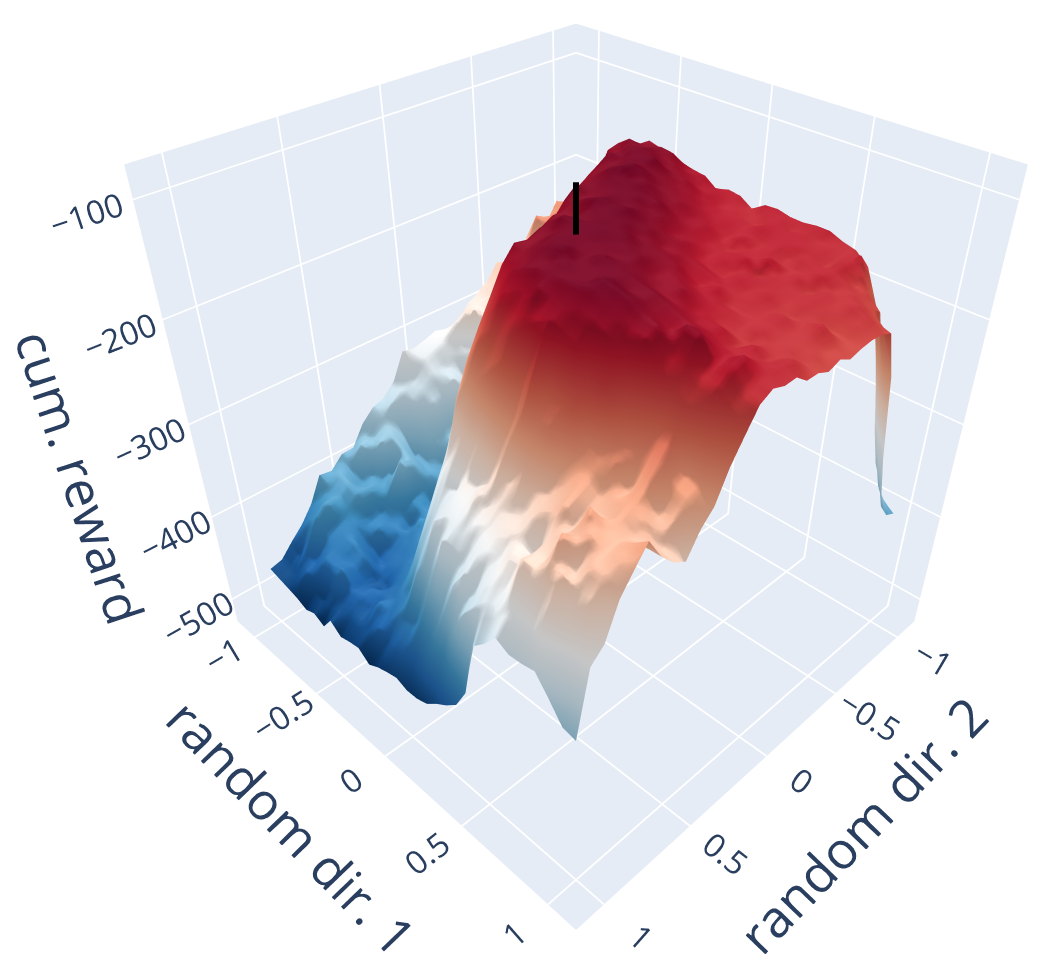}
        \caption{Position control, reward}
        \label{subfig:opt_vis_gym_pendulum_pc_reward}
    \end{subfigure}
    \begin{subfigure}[b]{\surfacefigwidthappendix}
        \centering
        \includegraphics[width=\textwidth]{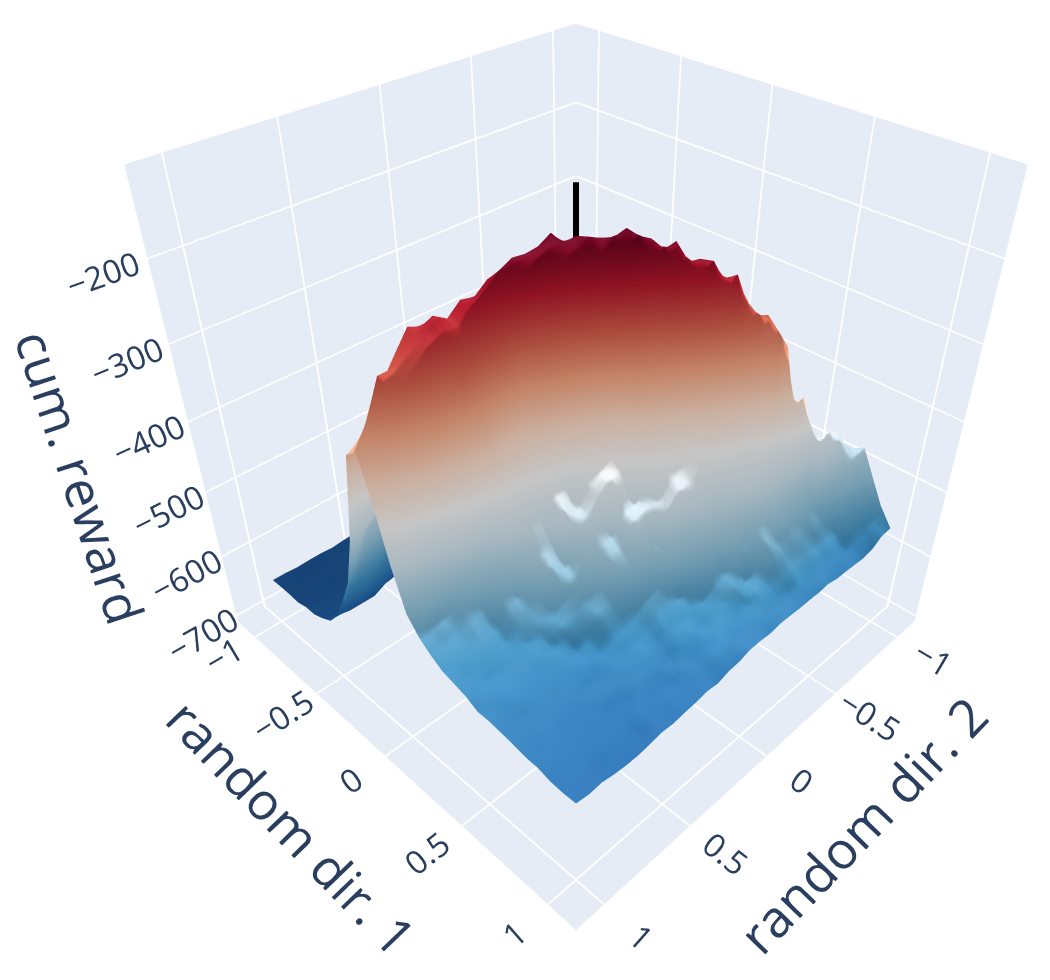}
        \caption{Velocity control, reward}
        \label{subfig:opt_vis_gym_pendulum_vc_reward}
    \end{subfigure} \\
    \begin{subfigure}[b]{\surfacefigwidthappendix}
        \centering
        \includegraphics[width=\textwidth]{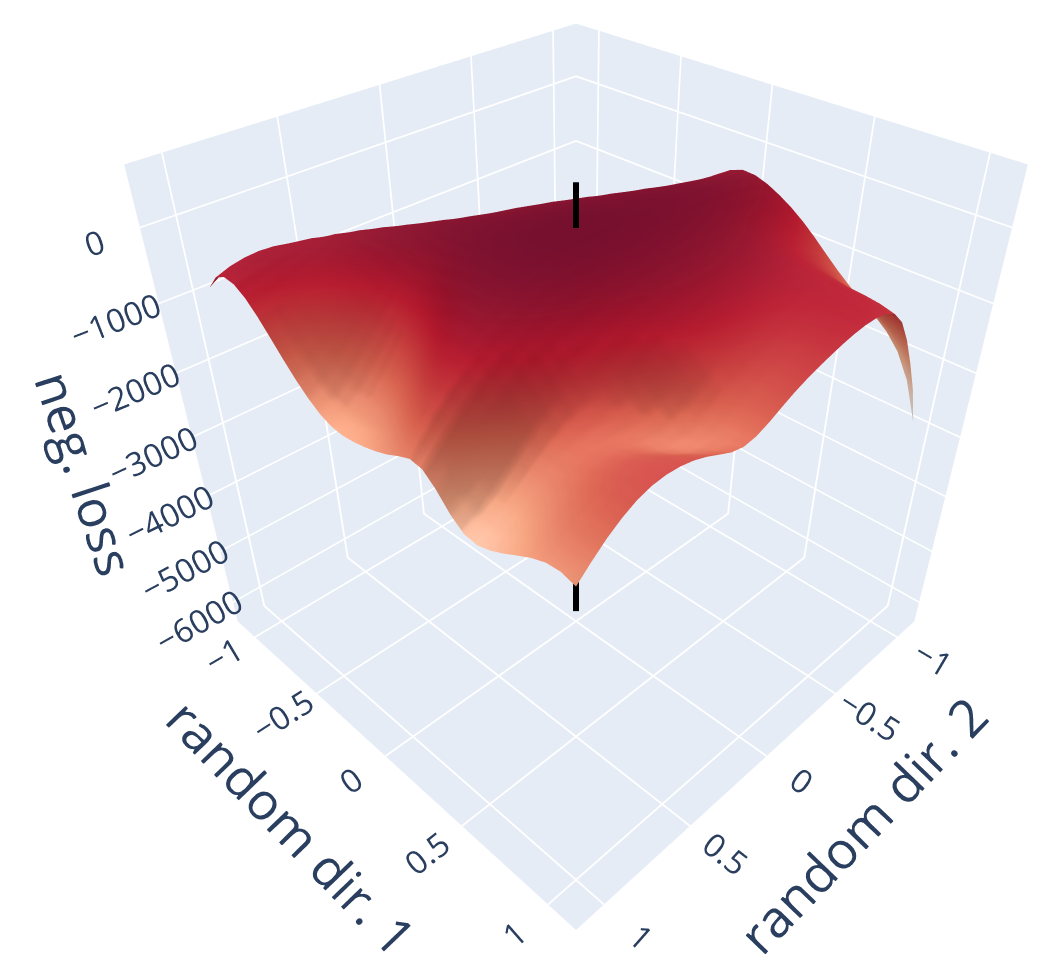}
        \caption{Torque control, negative loss}
        \label{subfig:opt_vis_gym_pendulum_tc_combined}
    \end{subfigure}
    \begin{subfigure}[b]{\surfacefigwidthappendix}
        \centering
        \includegraphics[width=\textwidth]{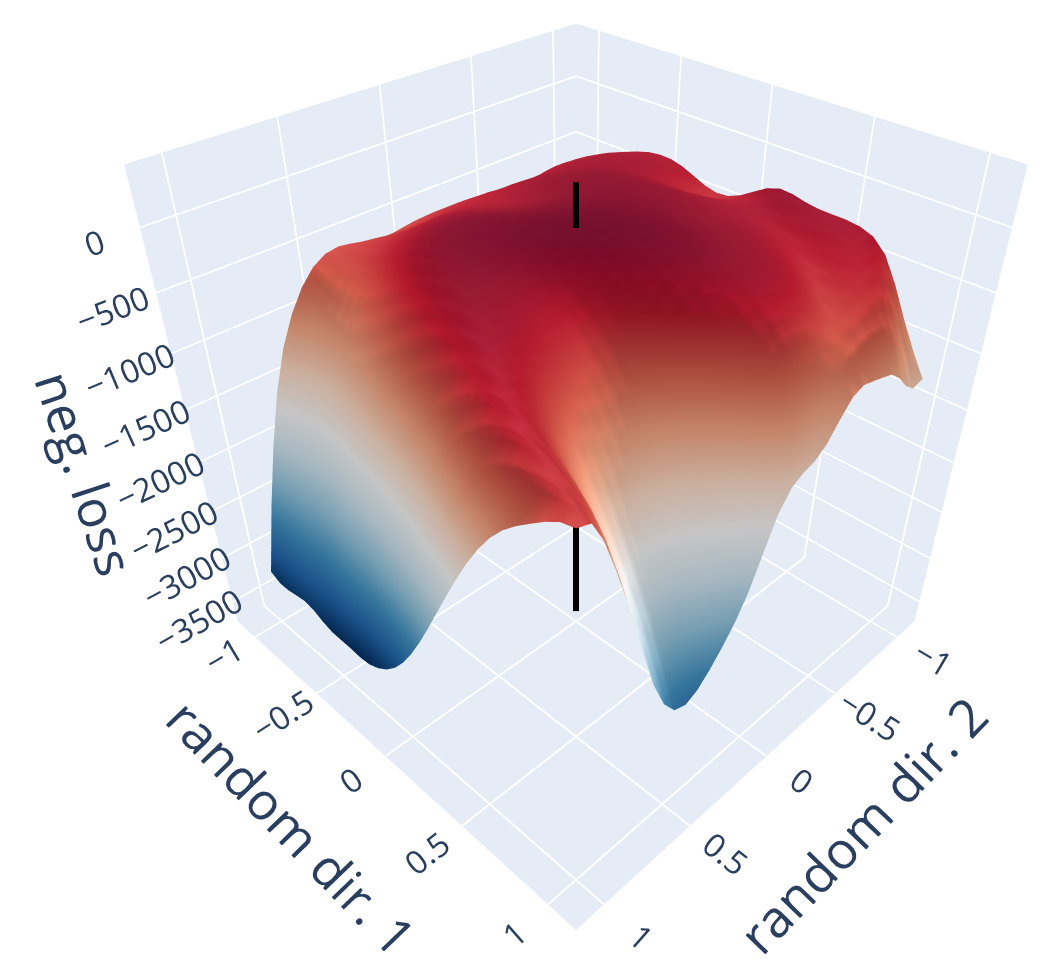}
        \caption{Position control, negative loss}
        \label{subfig:opt_vis_gym_pendulum_pc_combined}
    \end{subfigure}
    \begin{subfigure}[b]{\surfacefigwidthappendix}
        \centering
        \includegraphics[width=\textwidth]{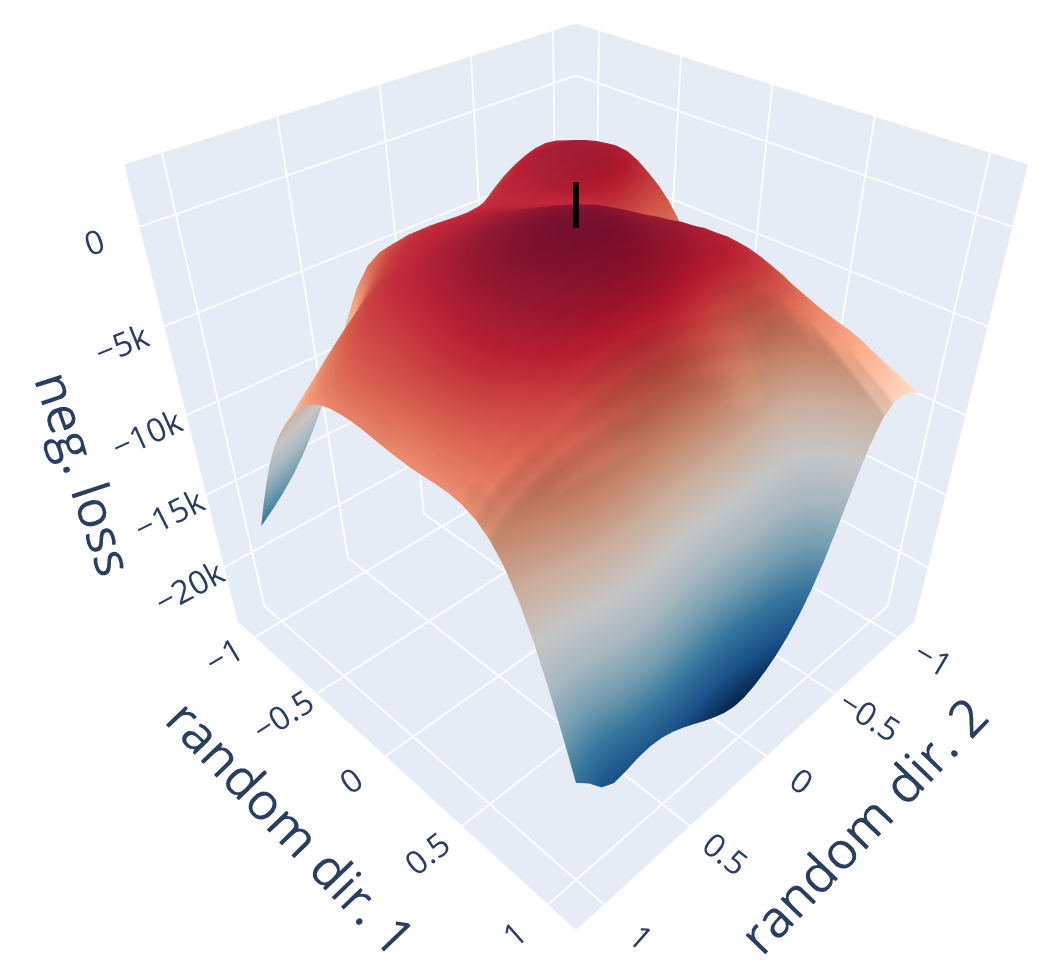}
        \caption{Velocity control, negative loss}
        \label{subfig:opt_vis_gym_pendulum_vc_combined}
    \end{subfigure}
    \caption{
        Visualizations of the optimization surface for the Pendulum task.
    }
    \label{fig:opt_vis_gym_pendulum}
\end{figure}

\begin{figure}[H]
    \centering
    \begin{subfigure}[b]{\surfacefigwidthappendix}
        \centering
        \includegraphics[width=\textwidth]{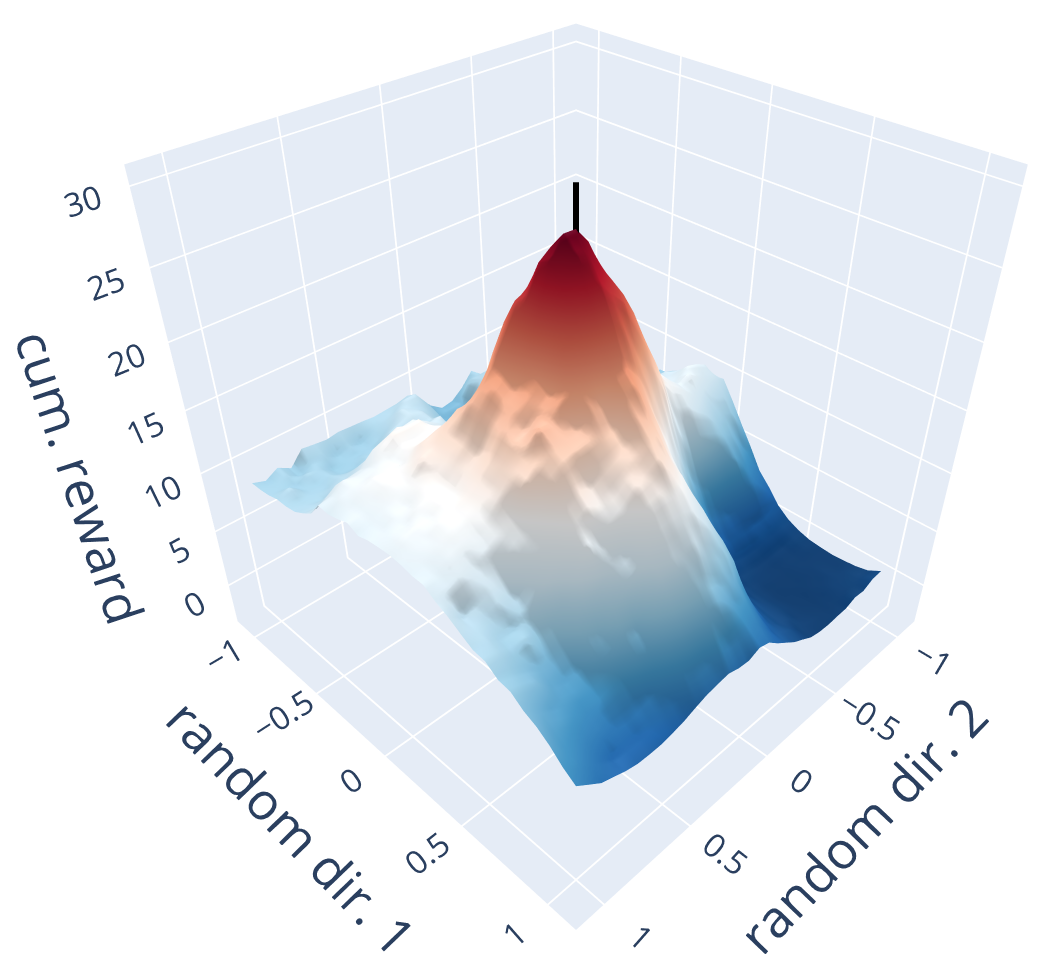}
        \caption{Torque control, reward}
        \label{subfig:opt_vis_dmc_cheetah_run_tc_reward}
    \end{subfigure}
    \begin{subfigure}[b]{\surfacefigwidthappendix}
        \centering
        \includegraphics[width=\textwidth]{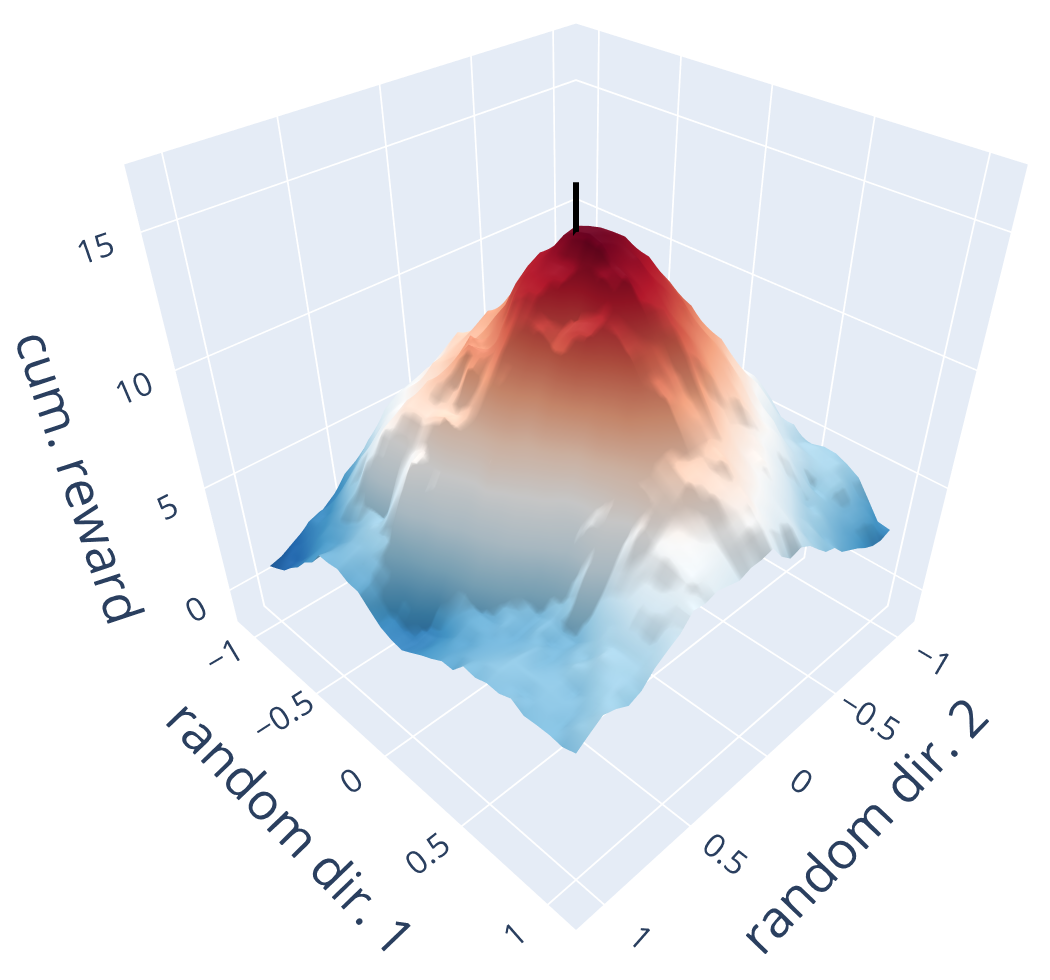}
        \caption{Position control, reward}
        \label{subfig:opt_vis_dmc_cheetah_run_pc_reward}
    \end{subfigure} \\
    \begin{subfigure}[b]{\surfacefigwidthappendix}
        \centering
        \includegraphics[width=\textwidth]{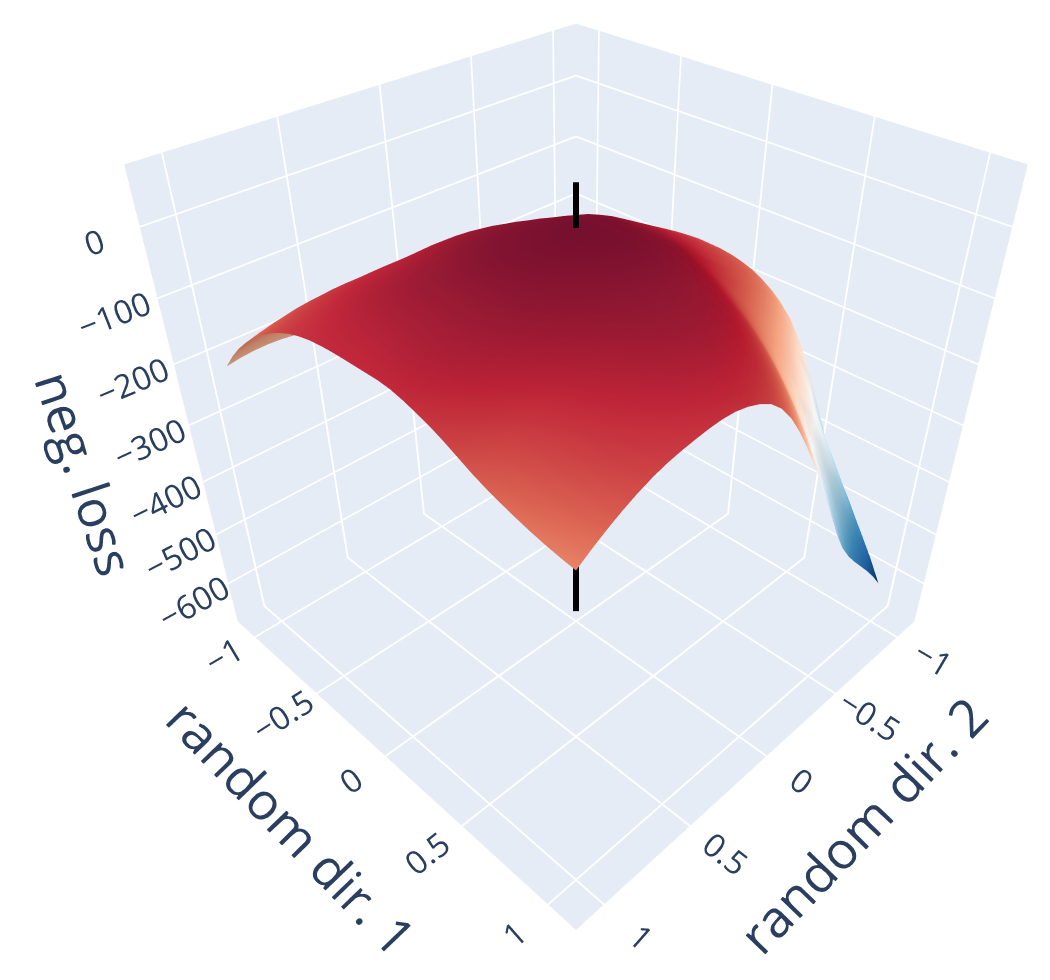}
        \caption{Torque control, negative loss}
        \label{subfig:opt_vis_dmc_cheetah_run_tc_combined}
    \end{subfigure}
    \begin{subfigure}[b]{\surfacefigwidthappendix}
        \centering
        \includegraphics[width=\textwidth]{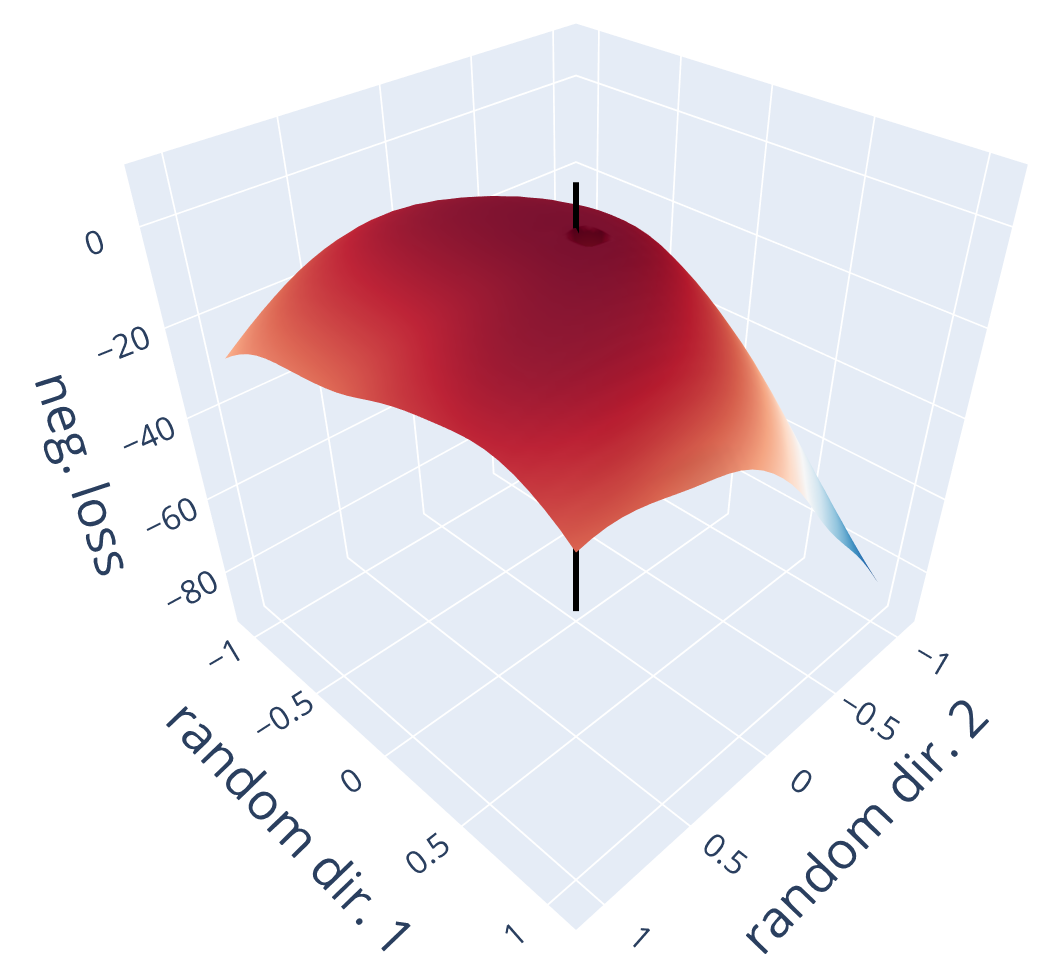}
        \caption{Position control, negative loss}
        \label{subfig:opt_vis_dmc_cheetah_run_pc_combined}
    \end{subfigure}
    \caption{
        Visualizations of the optimization surface for the Cheetah-run task.
    }
    \label{fig:opt_vis_dmc_cheetah_run}
\end{figure}

\begin{figure}[H]
    \centering
    \begin{subfigure}[b]{\surfacefigwidthappendix}
        \centering
        \includegraphics[width=\textwidth]{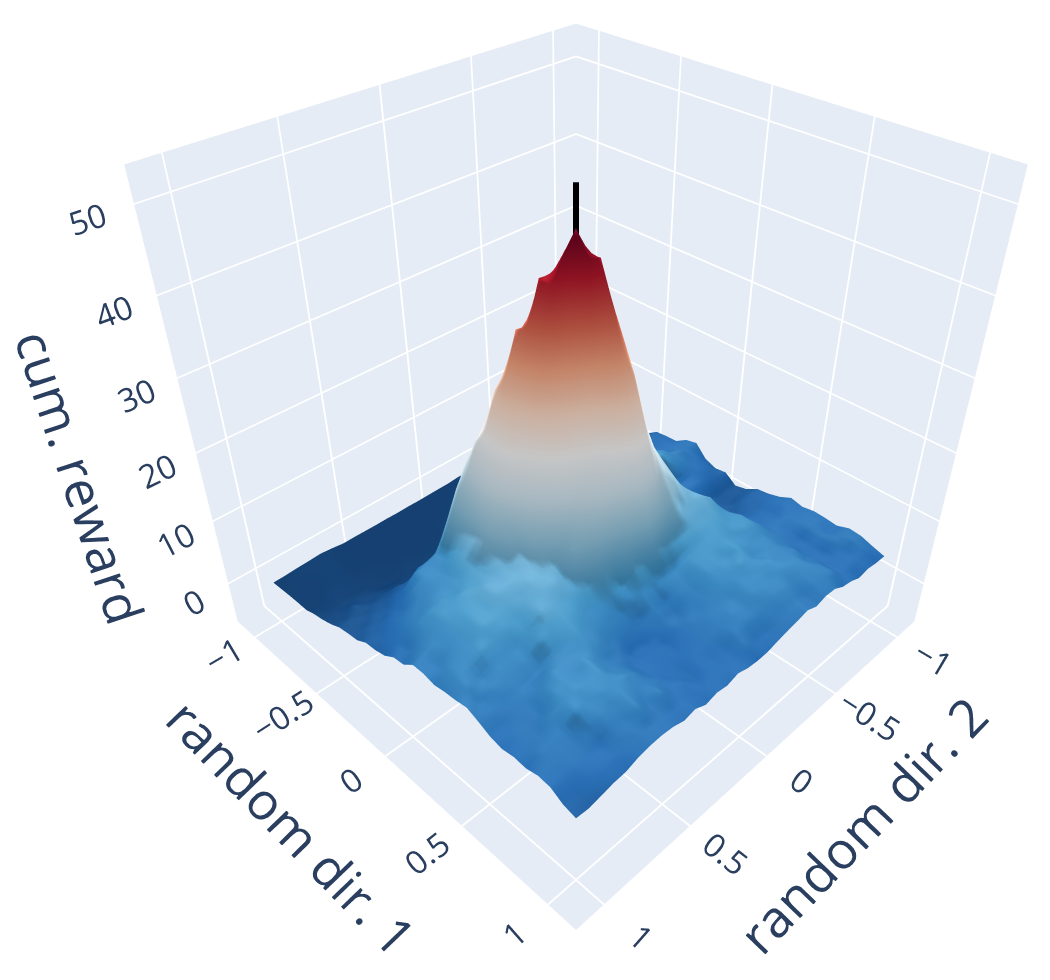}
        \caption{Torque control, reward}
        \label{subfig:opt_vis_dmc_finger_spin_tc_reward}
    \end{subfigure}
    \begin{subfigure}[b]{\surfacefigwidthappendix}
        \centering
        \includegraphics[width=\textwidth]{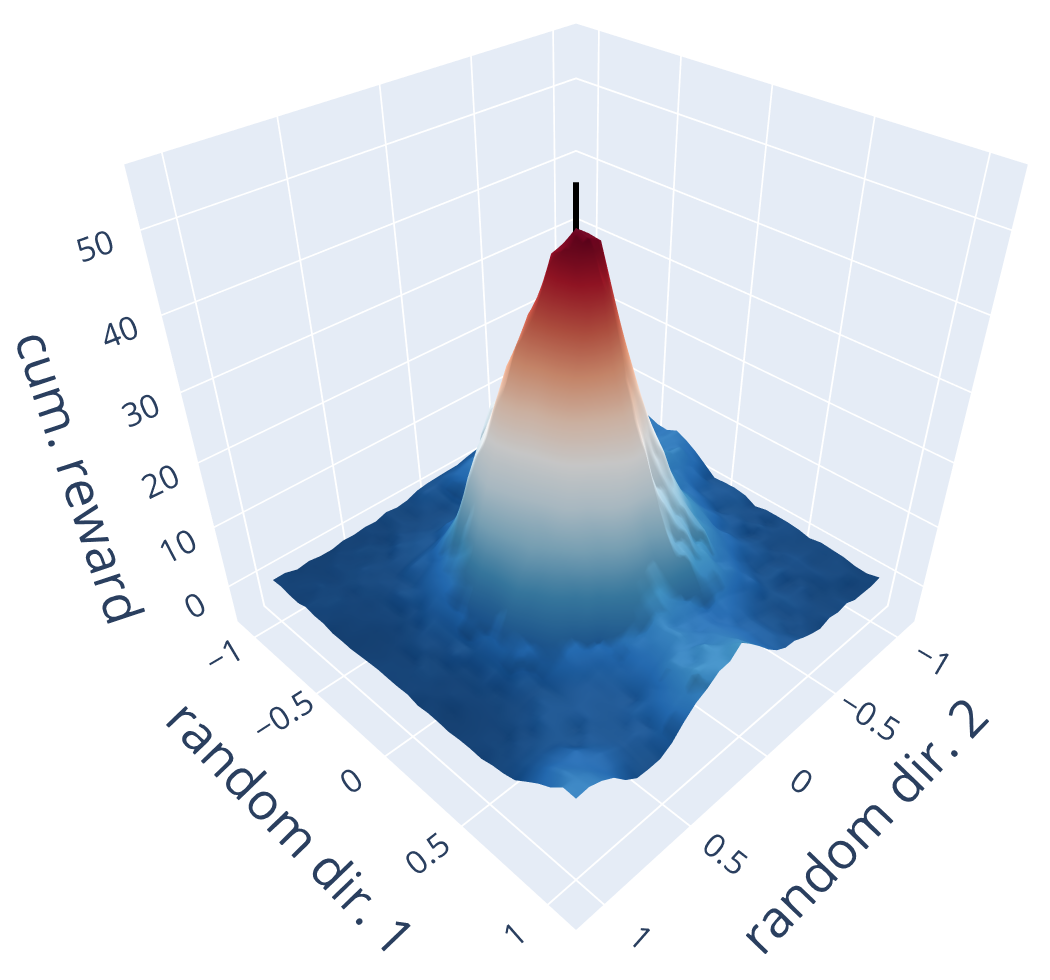}
        \caption{Position control, reward}
        \label{subfig:opt_vis_dmc_finger_spin_pc_reward}
    \end{subfigure} \\
    \begin{subfigure}[b]{\surfacefigwidthappendix}
        \centering
        \includegraphics[width=\textwidth]{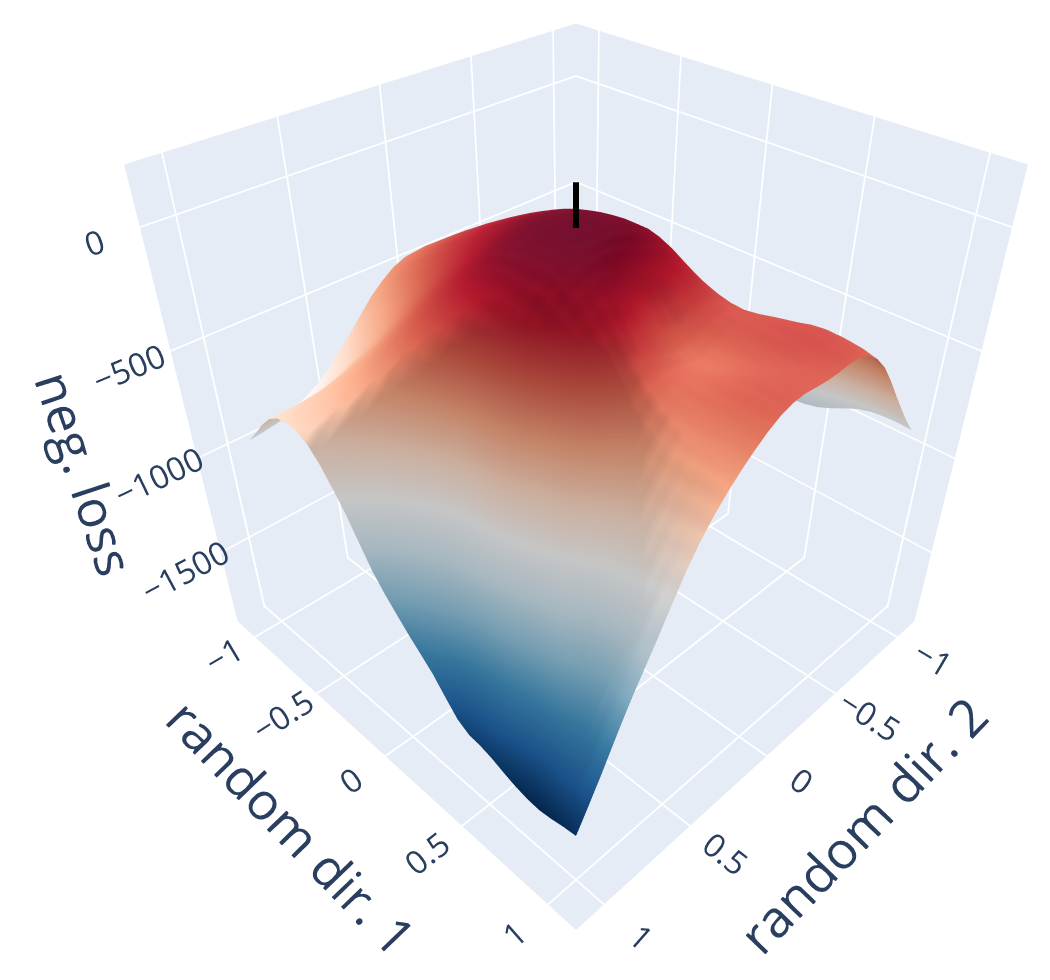}
        \caption{Torque control, negative loss}
        \label{subfig:opt_vis_dmc_finger_spin_tc_combined}
    \end{subfigure}
    \begin{subfigure}[b]{\surfacefigwidthappendix}
        \centering
        \includegraphics[width=\textwidth]{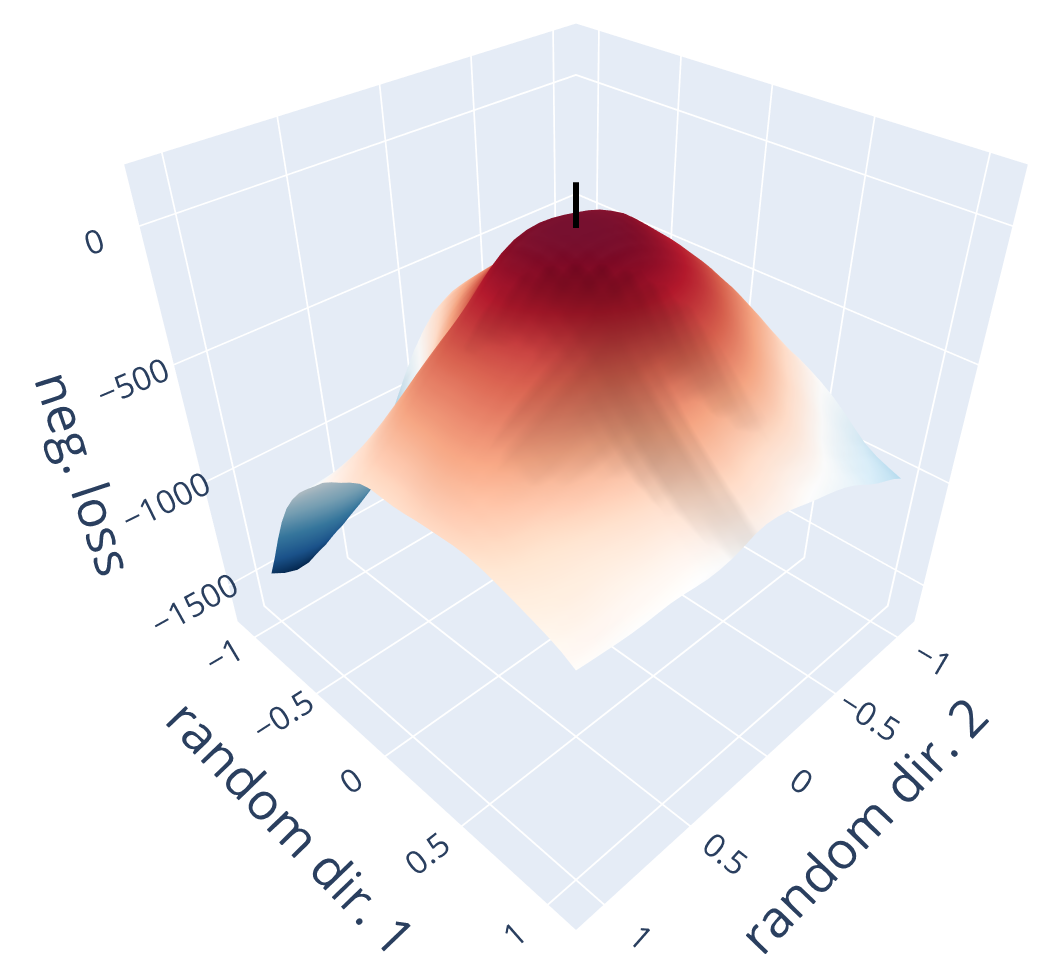}
        \caption{Position control, negative loss}
        \label{subfig:opt_vis_dmc_finger_spin_pc_combined}
    \end{subfigure}
    \caption{
        Visualizations of the optimization surface for the Finger-spin task.
    }
    \label{fig:opt_vis_dmc_finger_spin}
\end{figure}

\begin{figure}[H]
    \centering
    \begin{subfigure}[b]{\surfacefigwidthappendix}
        \centering
        \includegraphics[width=\textwidth]{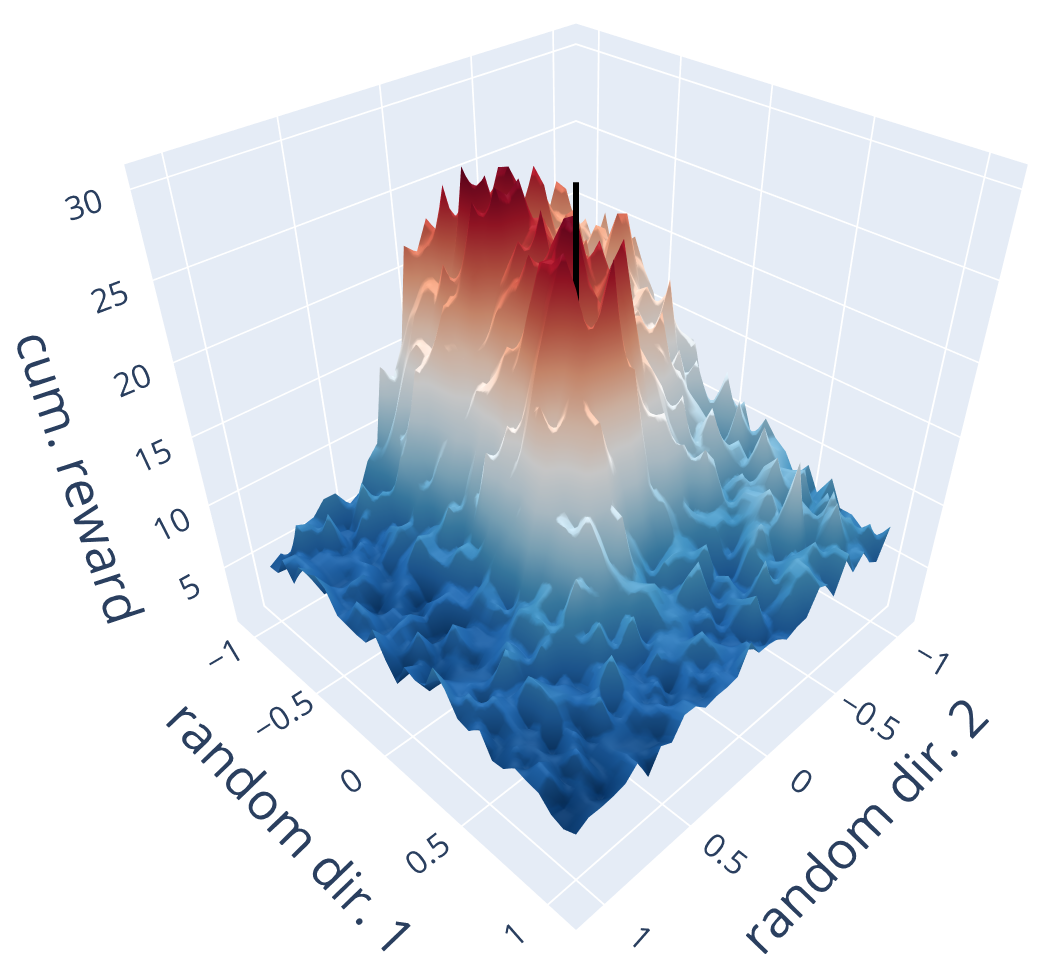}
        \caption{Torque control, reward}
        \label{subfig:opt_vis_dmc_reacher_easy_tc_reward}
    \end{subfigure}
    \begin{subfigure}[b]{\surfacefigwidthappendix}
        \centering
        \includegraphics[width=\textwidth]{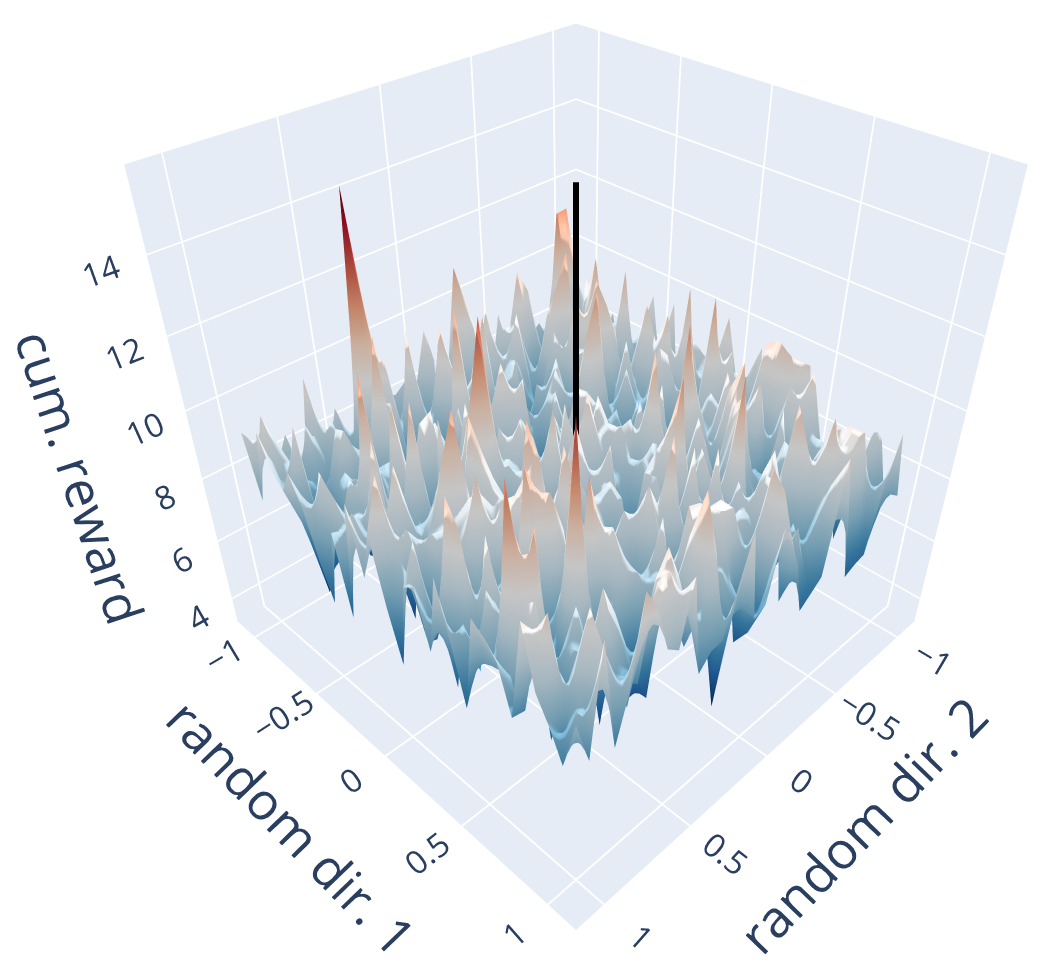}
        \caption{Position control, reward}
        \label{subfig:opt_vis_dmc_reacher_easy_pc_reward}
    \end{subfigure} \\
    \begin{subfigure}[b]{\surfacefigwidthappendix}
        \centering
        \includegraphics[width=\textwidth]{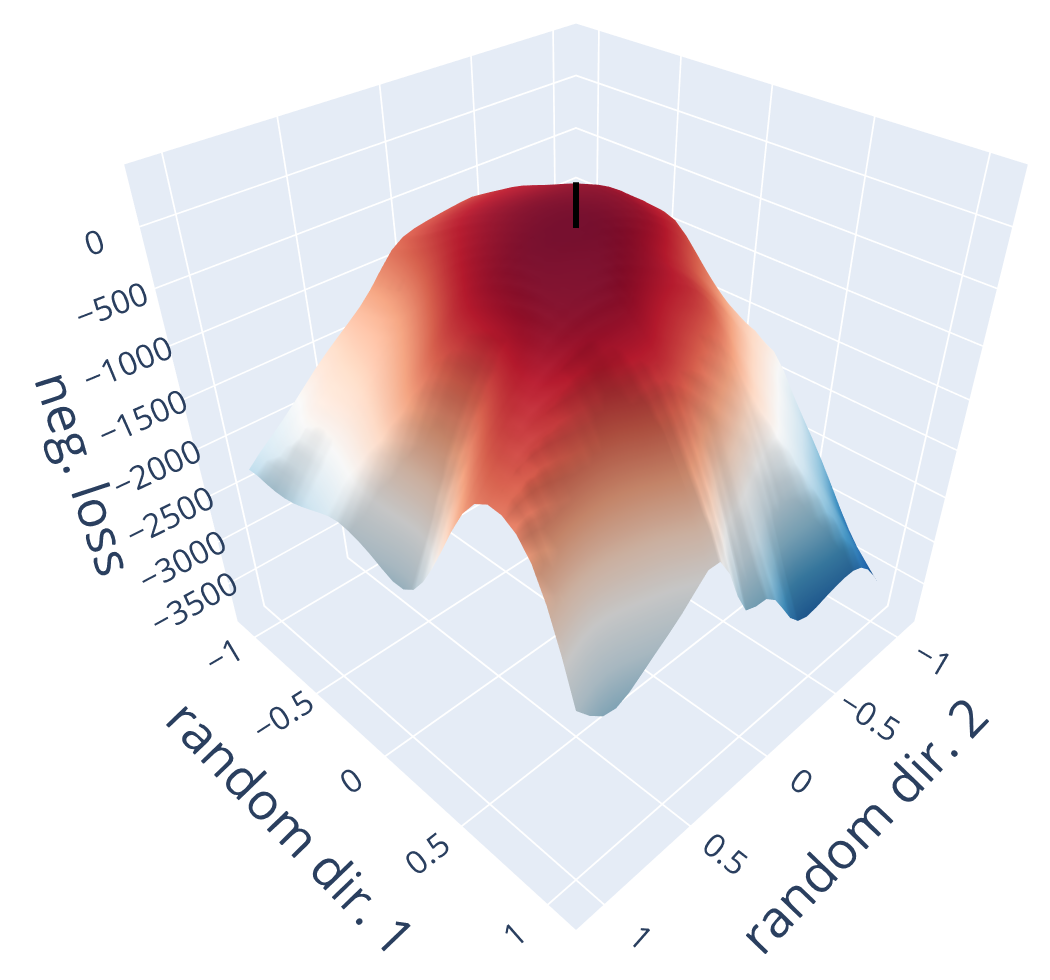}
        \caption{Torque control, negative loss}
        \label{subfig:opt_vis_dmc_reacher_easy_tc_combined}
    \end{subfigure}
    \begin{subfigure}[b]{\surfacefigwidthappendix}
        \centering
        \includegraphics[width=\textwidth]{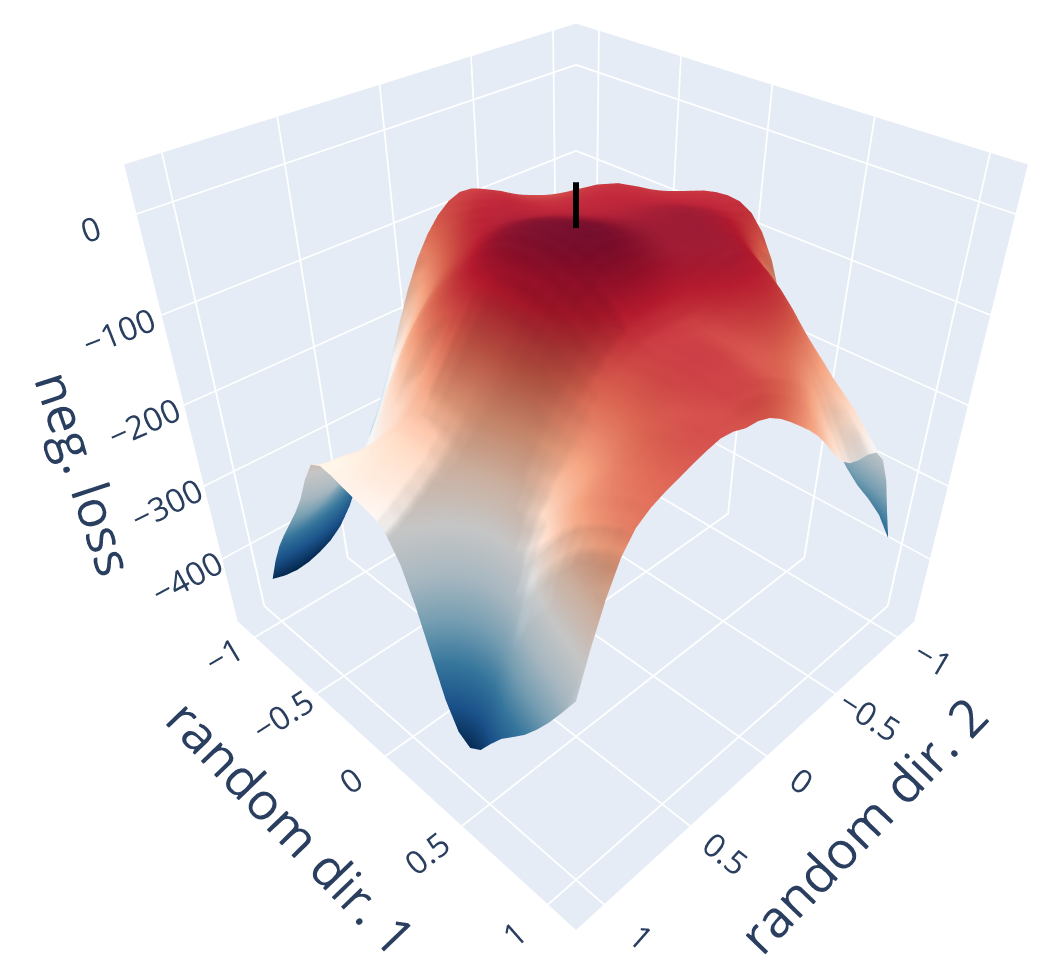}
        \caption{Position control, negative loss}
        \label{subfig:opt_vis_dmc_reacher_easy_pc_combined}
    \end{subfigure} \\
    \caption{
        Visualizations of the optimization surface for the Reacher-easy task.
    }
    \label{fig:opt_vis_dmc_reacher_easy}
\end{figure}

\pagebreak

\subsubsection{Accuracy of the gradient estimates}
\mbox{}     %
\begin{figure}[H]
    \centering
    \includegraphics[width=0.6\textwidth]{figures/legend.pdf} \\
    \vspace{0.2cm}
    \begin{subfigure}[b]{0.3\textwidth}
        \centering
        \includegraphics[width=\textwidth]{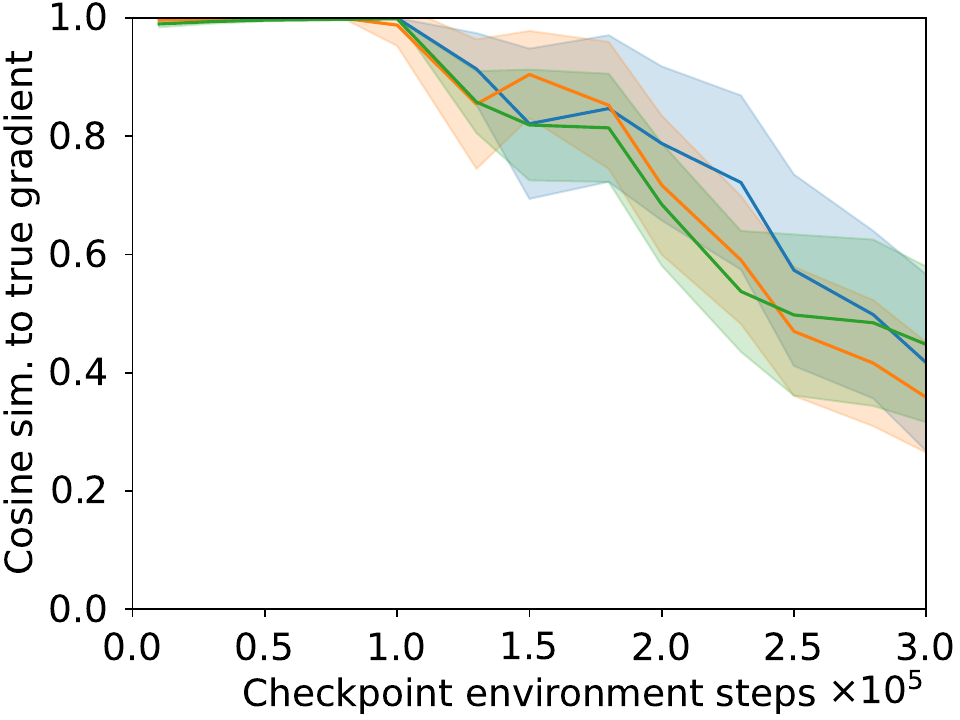}
        \caption{Pendulum, total loss}
        \label{subfig:gradient_sim_true_gym_pendulum_combined}
    \end{subfigure}
    \begin{subfigure}[b]{0.3\textwidth}
        \centering
        \includegraphics[width=\textwidth]{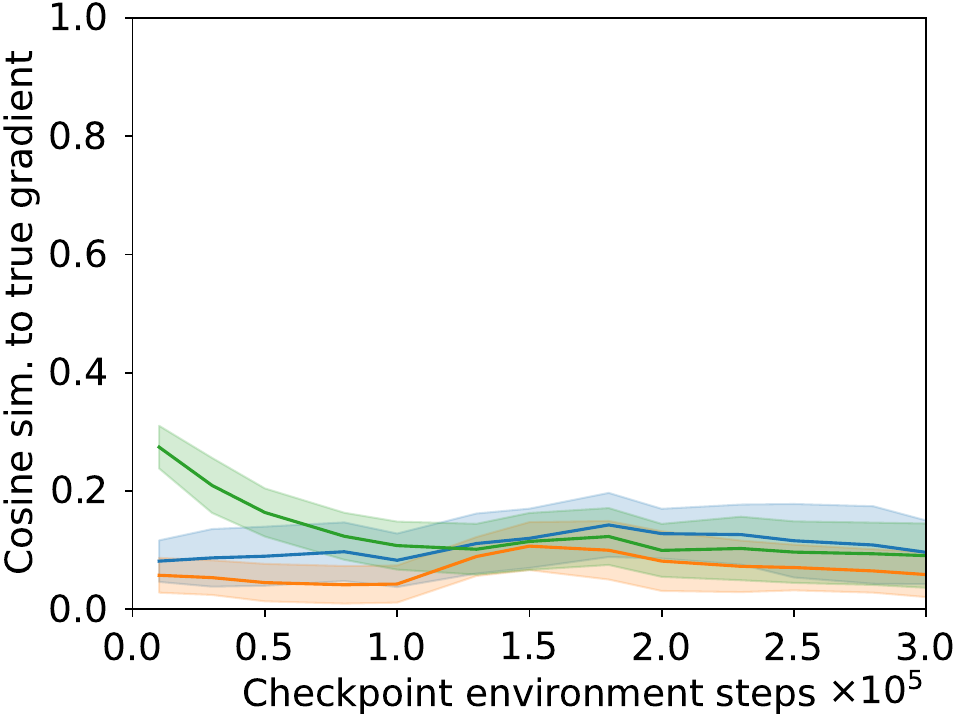}
        \caption{Pendulum, policy loss}
        \label{subfig:gradient_sim_true_gym_pendulum_policy}
    \end{subfigure}
    \begin{subfigure}[b]{0.3\textwidth}
        \centering
        \includegraphics[width=\textwidth]{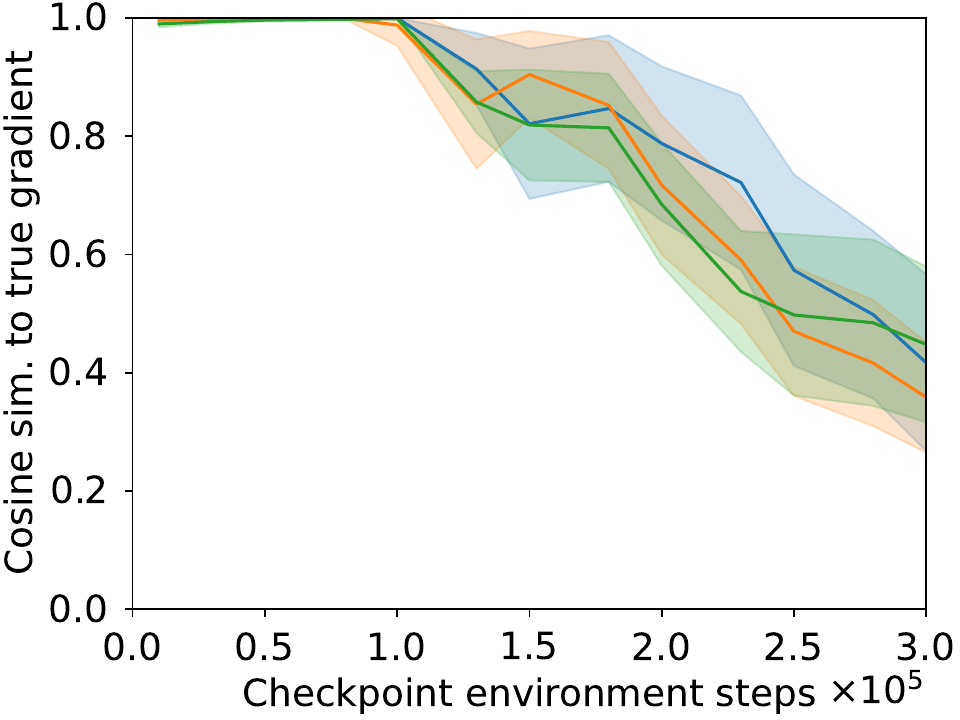}
        \caption{Pendulum, value function loss}
        \label{subfig:gradient_sim_true_gym_pendulum_vf}
    \end{subfigure}
    \begin{subfigure}[b]{0.3\textwidth}
        \centering
        \includegraphics[width=\textwidth]{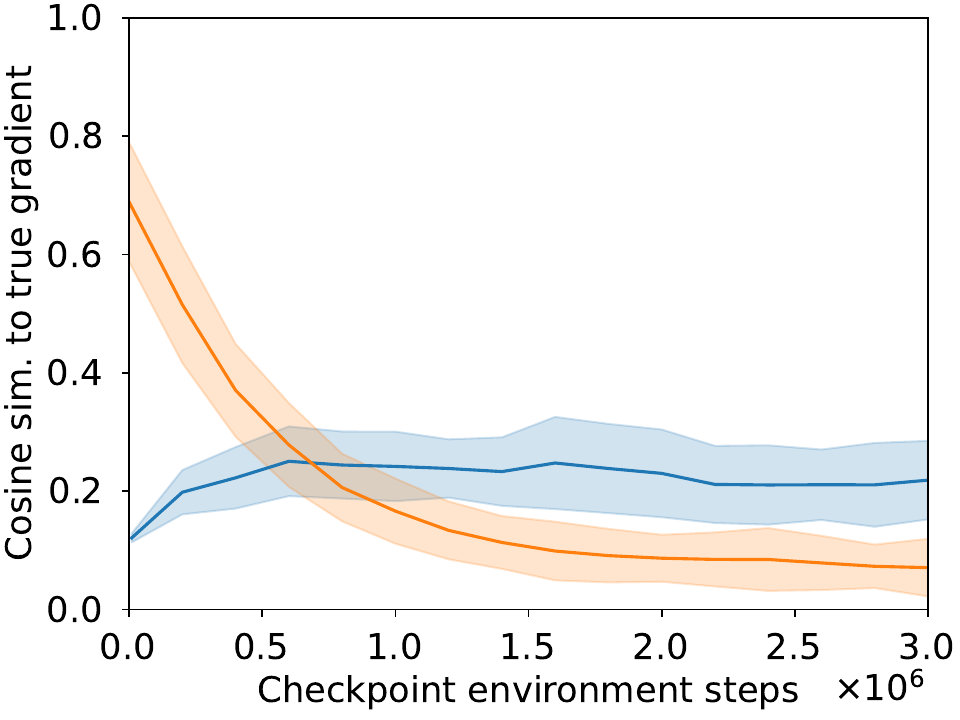}
        \caption{Cheetah-run, total loss}
        \label{subfig:gradient_sim_true_dmc_cheetah_run_combined}
    \end{subfigure}
    \begin{subfigure}[b]{0.3\textwidth}
        \centering
        \includegraphics[width=\textwidth]{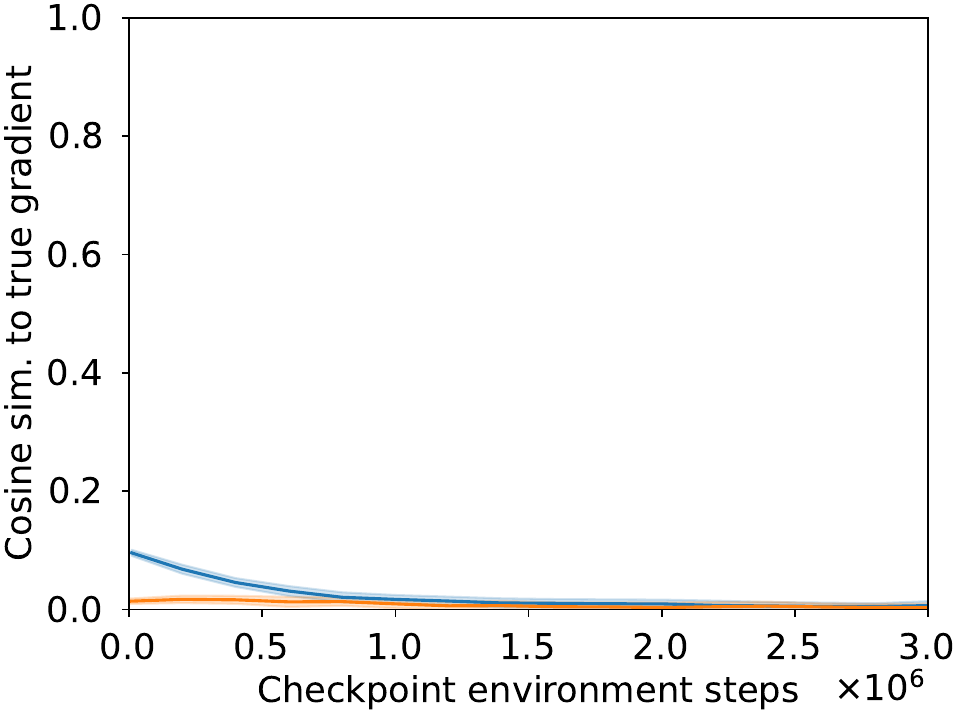}
        \caption{Cheetah-run, policy loss}
        \label{subfig:gradient_sim_true_dmc_cheetah_run_policy}
    \end{subfigure}
    \begin{subfigure}[b]{0.3\textwidth}
        \centering
        \includegraphics[width=\textwidth]{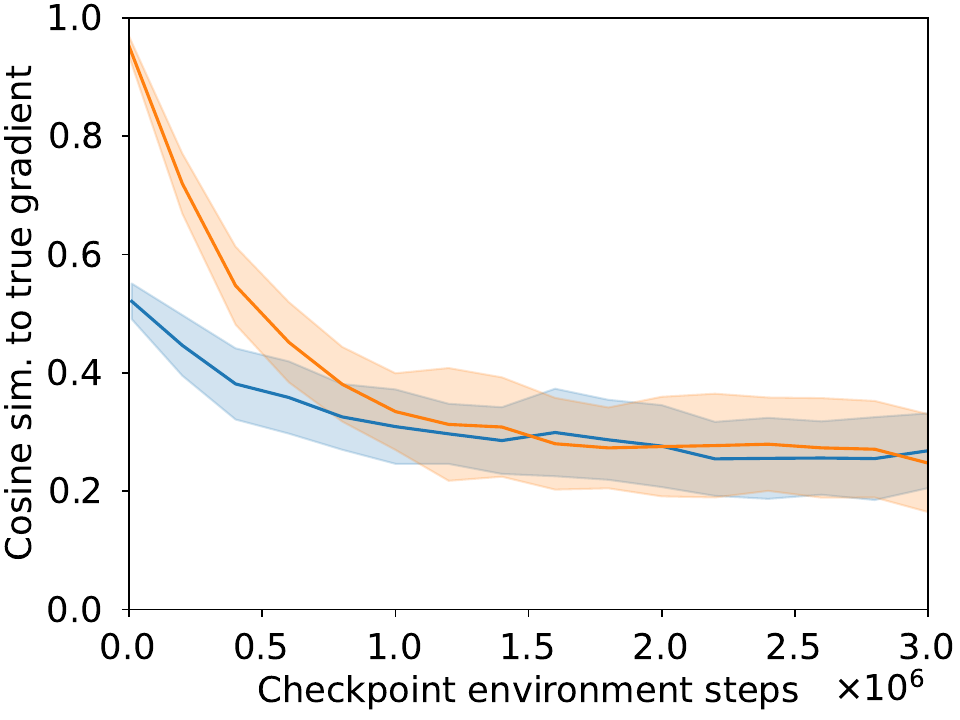}
        \caption{Cheetah-run, value function loss}
        \label{subfig:gradient_sim_true_dmc_cheetah_run_vf}
    \end{subfigure}
    \begin{subfigure}[b]{0.3\textwidth}
        \centering
        \includegraphics[width=\textwidth]{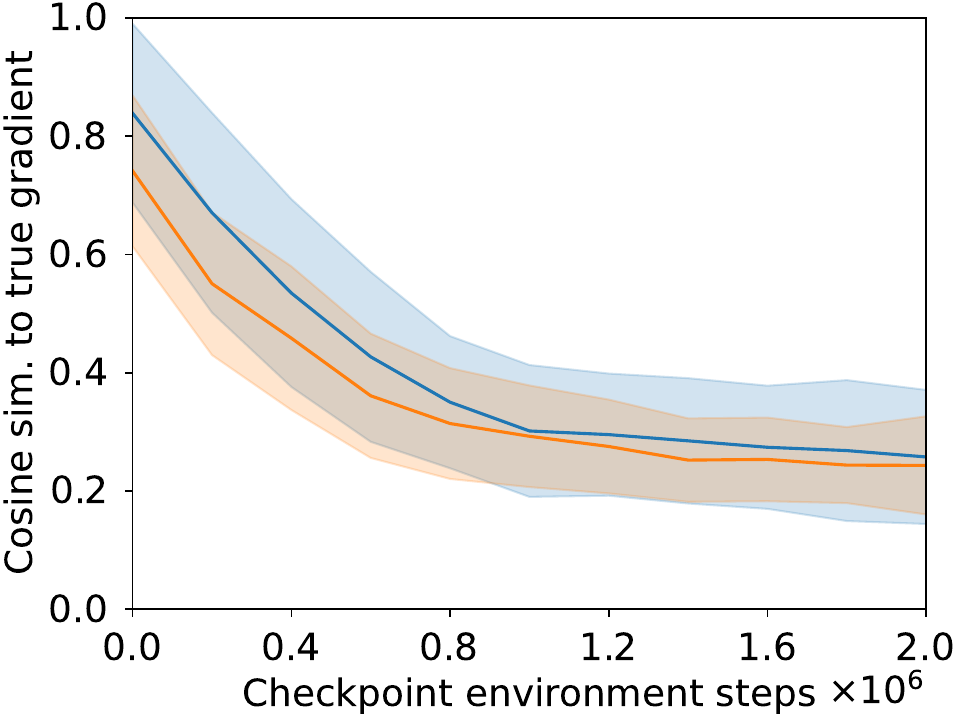}
        \caption{Finger-spin, total loss}
        \label{subfig:gradient_sim_true_dmc_finger_spin_combined}
    \end{subfigure}
    \begin{subfigure}[b]{0.3\textwidth}
        \centering
        \includegraphics[width=\textwidth]{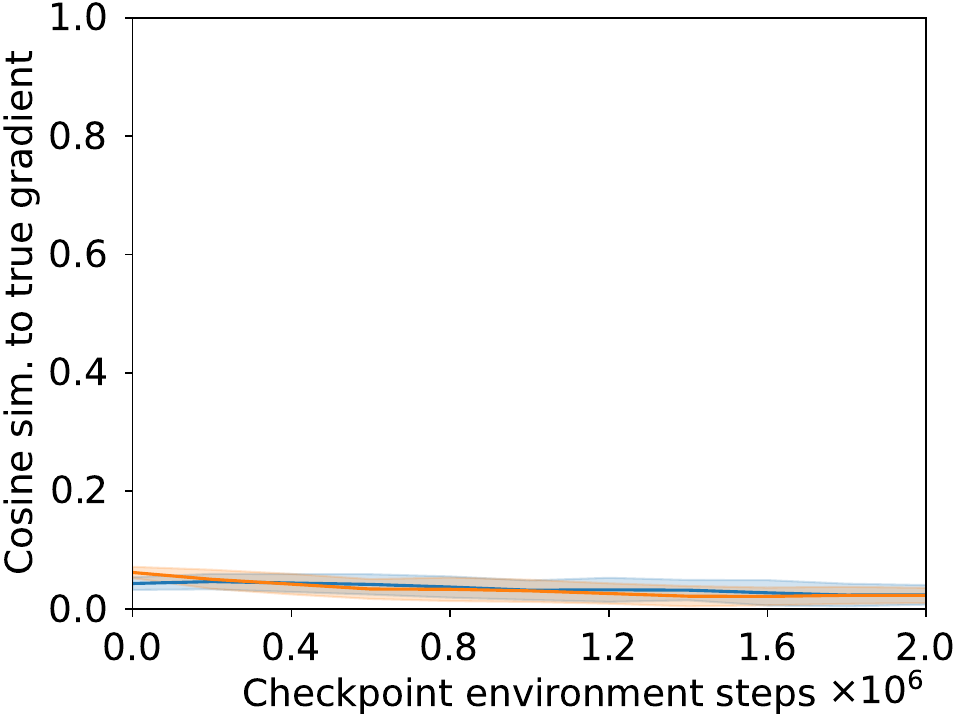}
        \caption{Finger-spin, policy loss}
        \label{subfig:gradient_sim_true_dmc_finger_spin_policy}
    \end{subfigure}
    \begin{subfigure}[b]{0.3\textwidth}
        \centering
        \includegraphics[width=\textwidth]{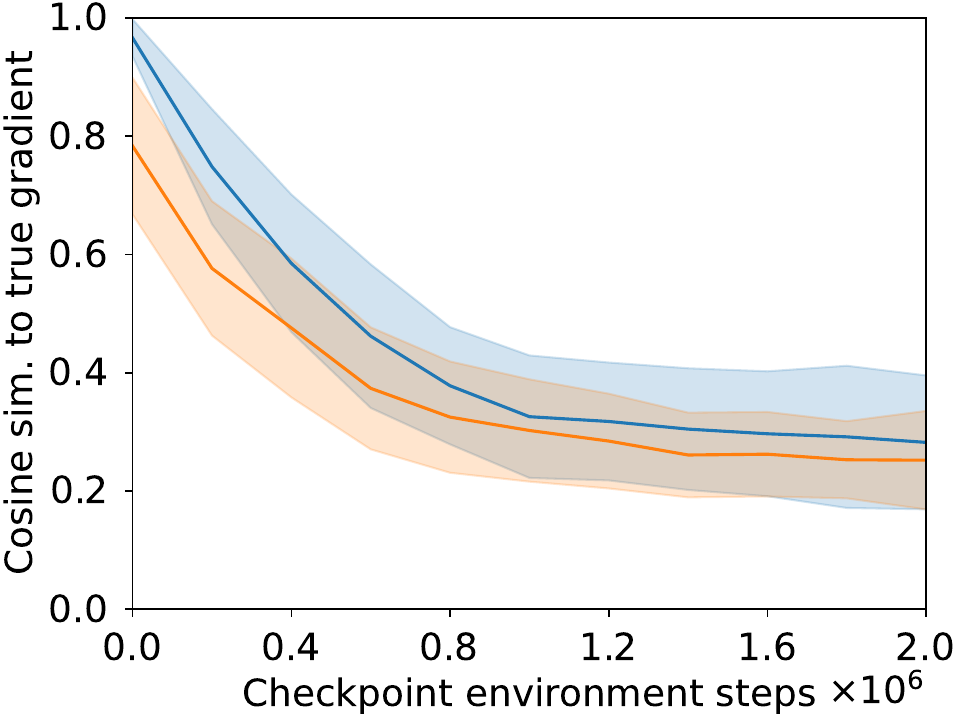}
        \caption{Finger-spin, value function loss}
        \label{subfig:gradient_sim_true_dmc_finger_spin_vf}
    \end{subfigure}
    \begin{subfigure}[b]{0.3\textwidth}
        \centering
        \includegraphics[width=\textwidth]{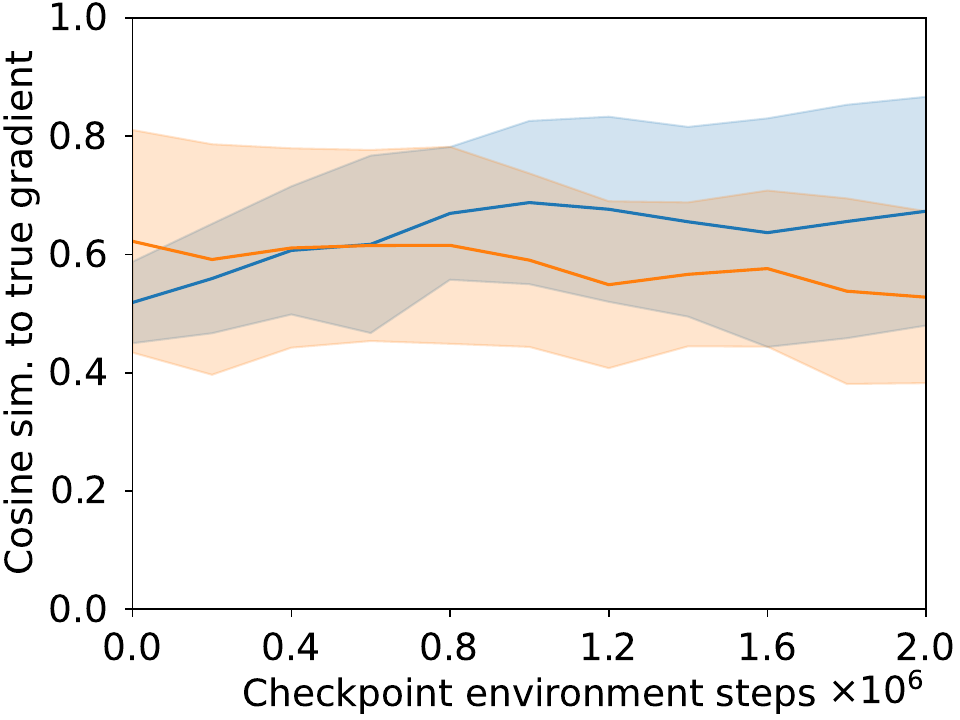}
        \caption{Reacher-easy, total loss}
        \label{subfig:gradient_sim_true_dmc_reacher_easy_combined}
    \end{subfigure}
    \hspace{0.005\textwidth}
    \begin{subfigure}[b]{0.3\textwidth}
        \centering
        \includegraphics[width=\textwidth]{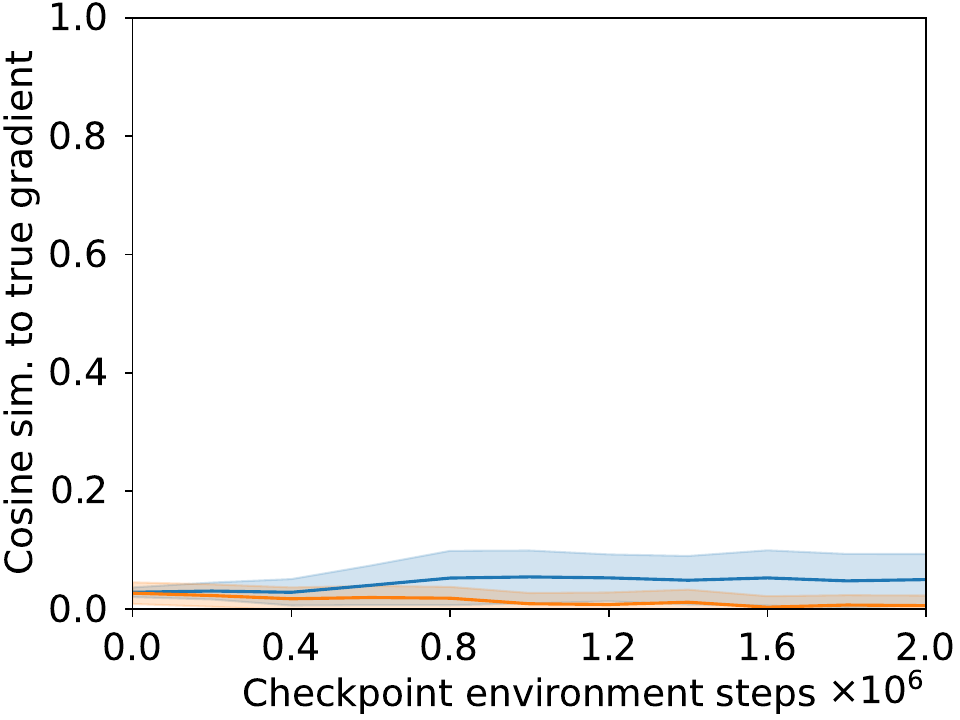}
        \caption{Reacher-easy, policy loss}
        \label{subfig:gradient_sim_true_dmc_reacher_easy_policy}
    \end{subfigure}
    \begin{subfigure}[b]{0.3\textwidth}
        \centering
        \includegraphics[width=\textwidth]{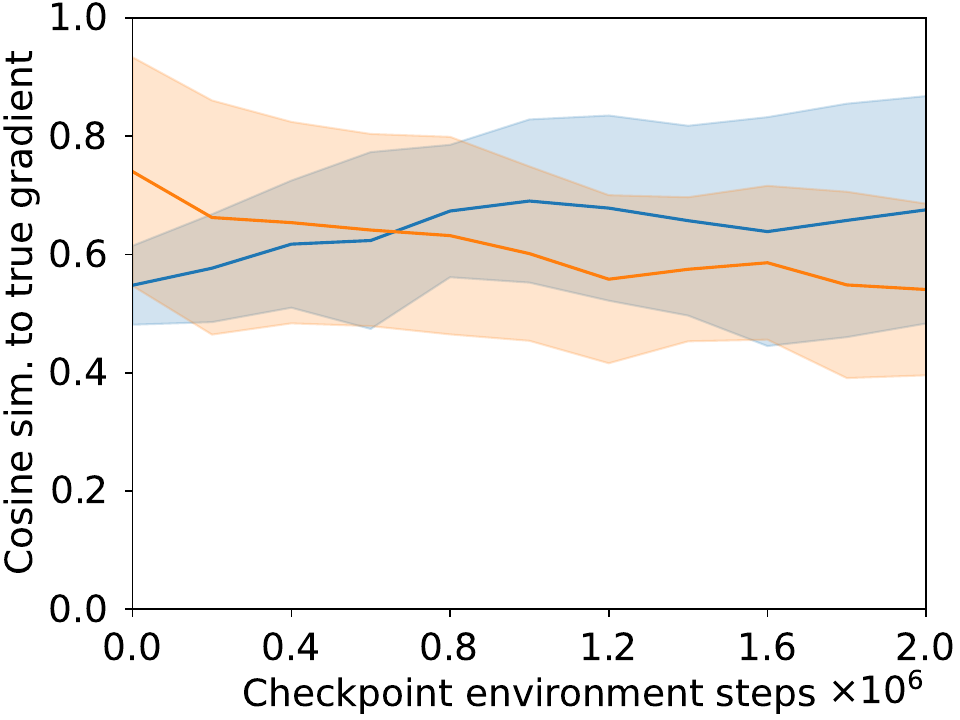}
        \caption{Reacher-easy, value function loss}
        \label{subfig:gradient_sim_true_dmc_reacher_easy_vf}
    \end{subfigure}
    \caption{
        The quality of the gradient estimates measured as the average cosine similarity to a good approximation of the true gradient, computed with $10^7$ samples.
        All plots are averaged over 10 training runs with different random seeds with the shaded region depicting the standard deviation across the training runs.
    }
    \label{fig:gradient_sim_true_other_tasks}
\end{figure}

\hspace{\baselineskip}

\subsection{Reproducibility of the optimization surface visualizations}
\label{app:opt_vis_reproducibility}

\begin{figure}[H]
    \centering
    \centering
    \begin{subfigure}[b]{\surfacefigwidthappendix}
        \includegraphics[width=\textwidth]{figures/optimization_surface_analysis/gym_reacher/gym_reacher_tc_random_directions_reward.png}
        \caption{Torque control, seed: 0}
    \end{subfigure}
    \begin{subfigure}[b]{\surfacefigwidthappendix}
        \includegraphics[width=\textwidth]{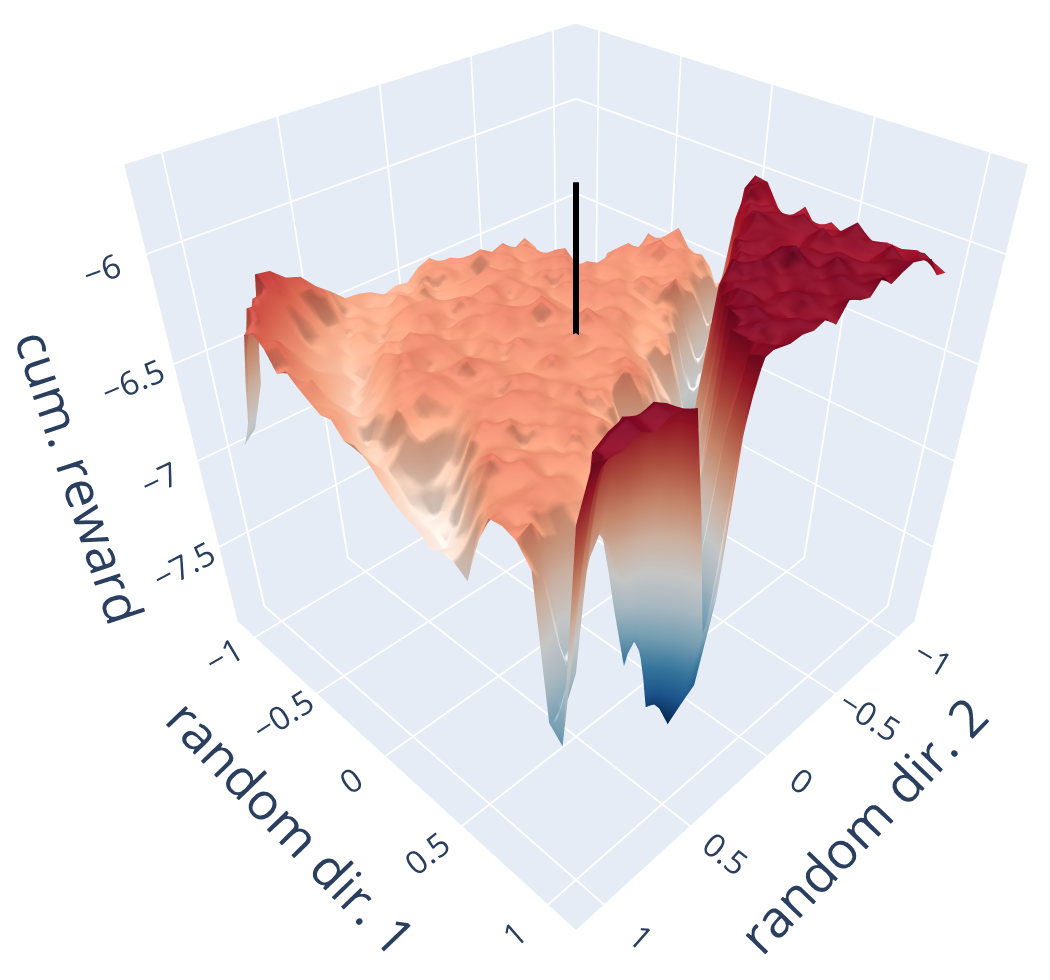}
        \caption{Torque control, seed: 1}
    \end{subfigure}
    \begin{subfigure}[b]{\surfacefigwidthappendix}
        \includegraphics[width=\textwidth]{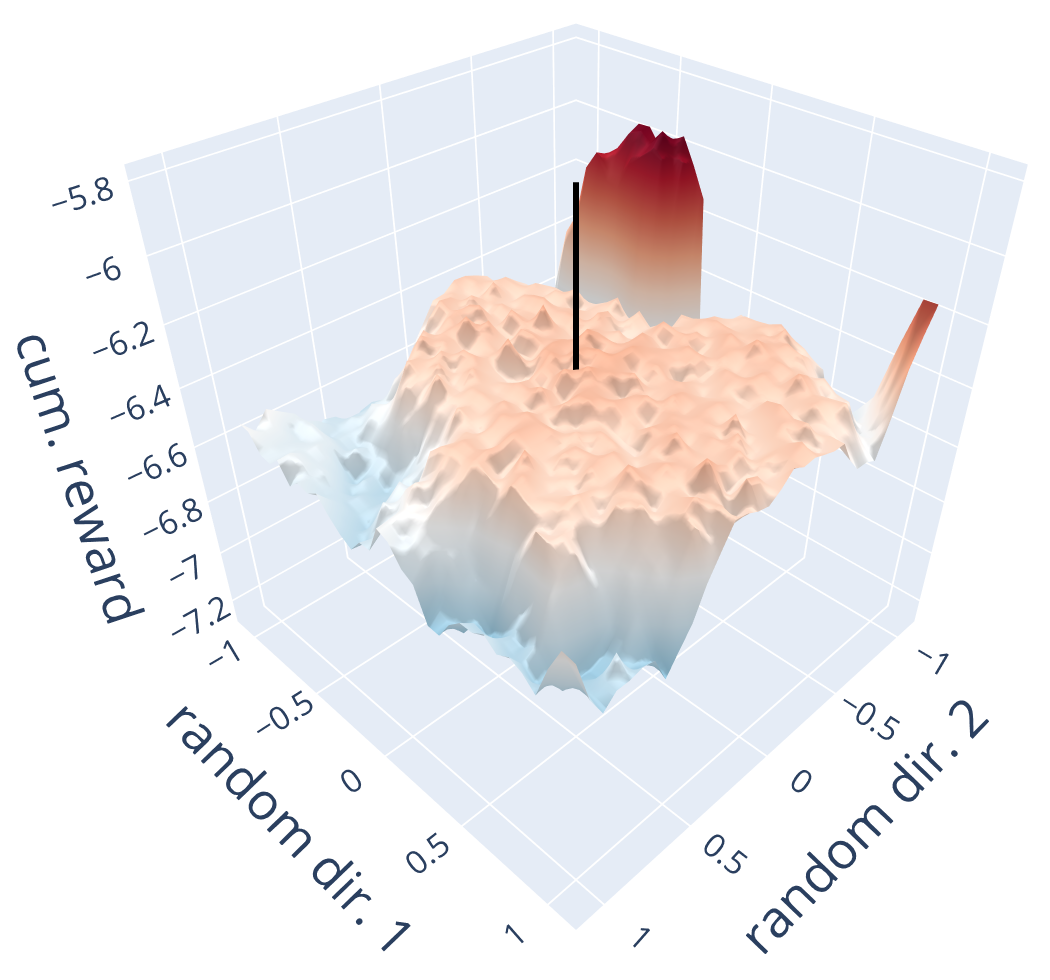}            
        \caption{Torque control, seed: 2}
    \end{subfigure} \\
        \begin{subfigure}[b]{\surfacefigwidthappendix}
        \includegraphics[width=\textwidth]{figures/optimization_surface_analysis/gym_reacher/gym_reacher_pc_random_directions_reward.png}
        \caption{Position control, seed: 0}
    \end{subfigure}
    \begin{subfigure}[b]{\surfacefigwidthappendix}
        \includegraphics[width=\textwidth]{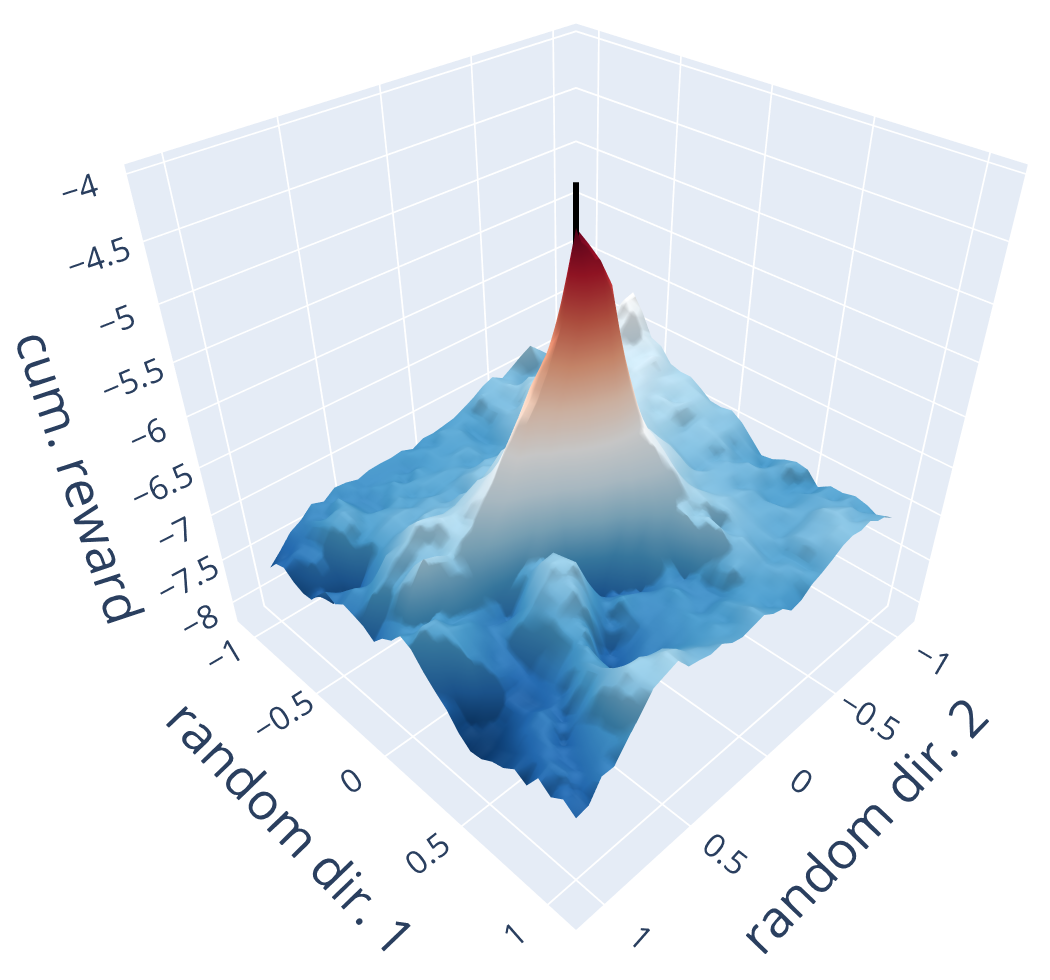}
        \caption{Position control, seed: 1}
    \end{subfigure}
    \begin{subfigure}[b]{\surfacefigwidthappendix}
        \includegraphics[width=\textwidth]{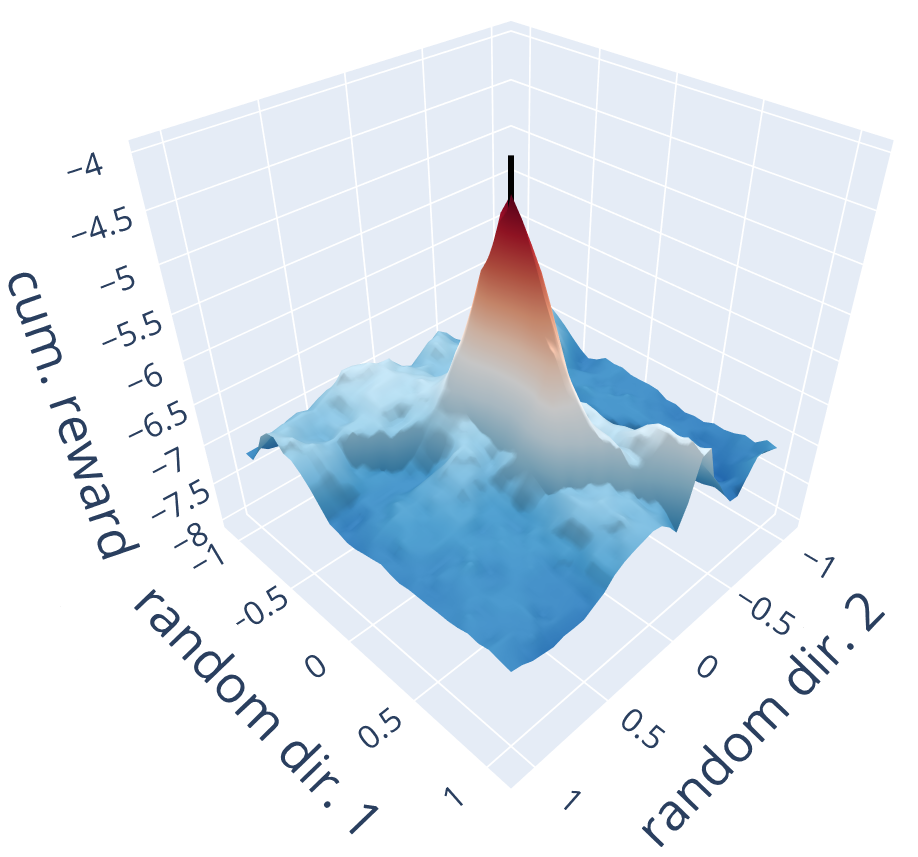} %
        \caption{Position control, seed: 2}
    \end{subfigure} \\
        \begin{subfigure}[b]{\surfacefigwidthappendix}
        \includegraphics[width=\textwidth]{figures/optimization_surface_analysis/gym_reacher/gym_reacher_vc_random_directions_reward.png}
        \caption{Velocity control, seed: 0}
    \end{subfigure}
    \begin{subfigure}[b]{\surfacefigwidthappendix}
        \includegraphics[width=\textwidth]{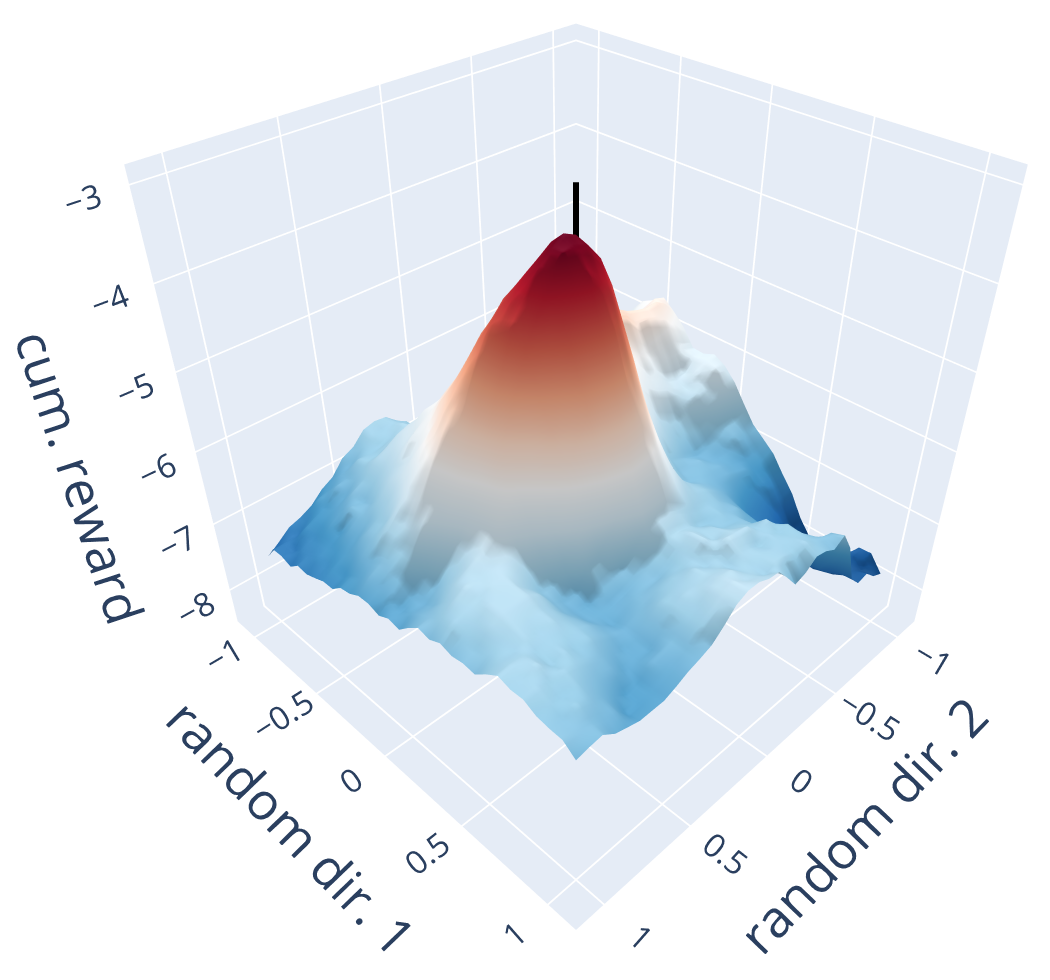}
        \caption{Velocity control, seed: 1}
    \end{subfigure}
    \begin{subfigure}[b]{\surfacefigwidthappendix}
        \includegraphics[width=\textwidth]{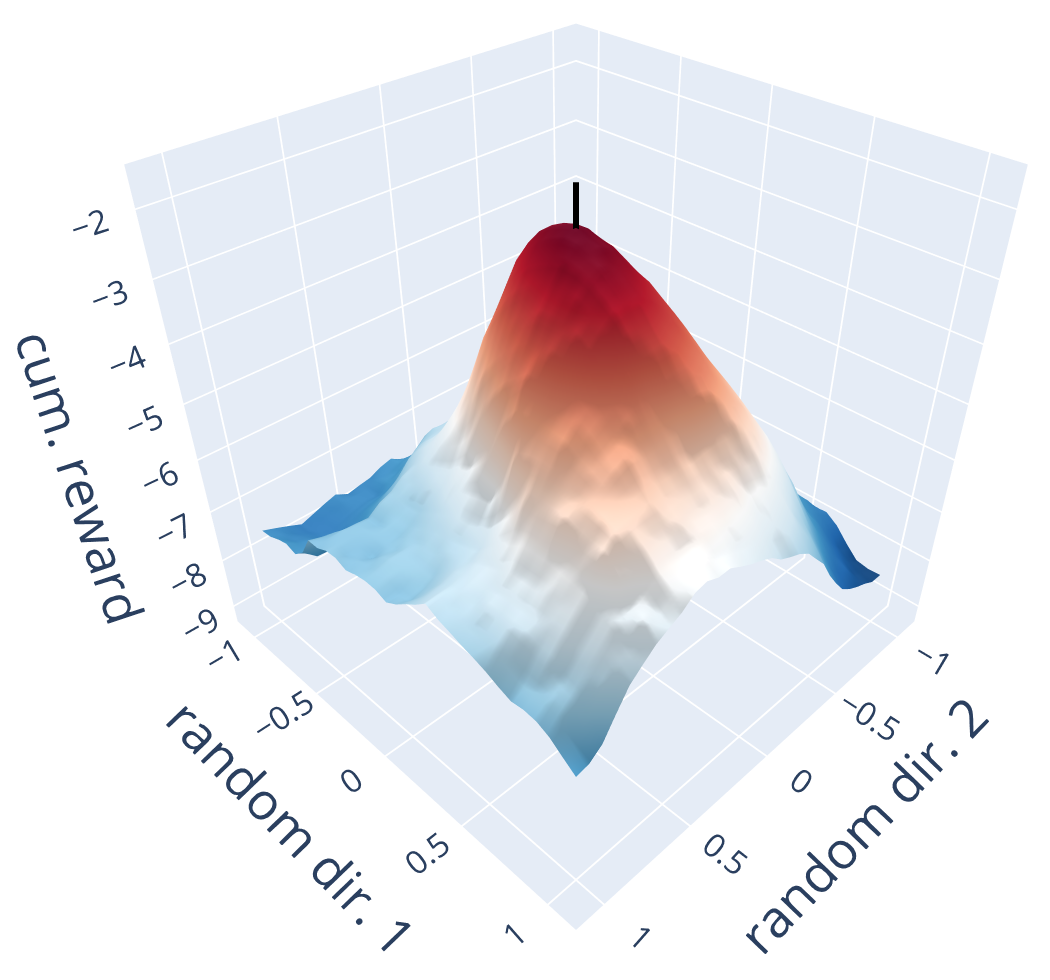}            
        \caption{Velocity control, seed: 2}
    \end{subfigure}
    \caption{
        Reproducibility of the reward surface plots on the Reacher task across training runs with different random seeds.
        All plots are generated after training for 1 million environment steps.
        The rows display results for different action representations.
        Across the columns, the representation is the same but the random seed used for training the agent and generating the random directions varies.
        Even though the visualization process includes a significant amount of stochasticity, we find that the plots for each action space configuration display similar characteristics.
    }
    \label{fig:opt_vis_reacher_repro}
\end{figure}

\pagebreak
\subsection{Training with more accurate gradient estimates}
\label{app:gt_gradient_training}

\begin{figure}[H]
    \centering
    \vspace{\baselineskip}
    \includegraphics[width=0.9\textwidth]{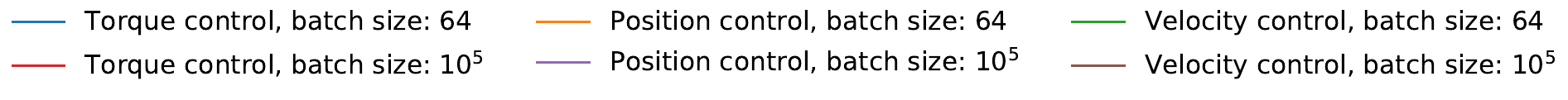} \\
    \vspace{0.2cm}
    \includegraphics[width=0.55\textwidth]{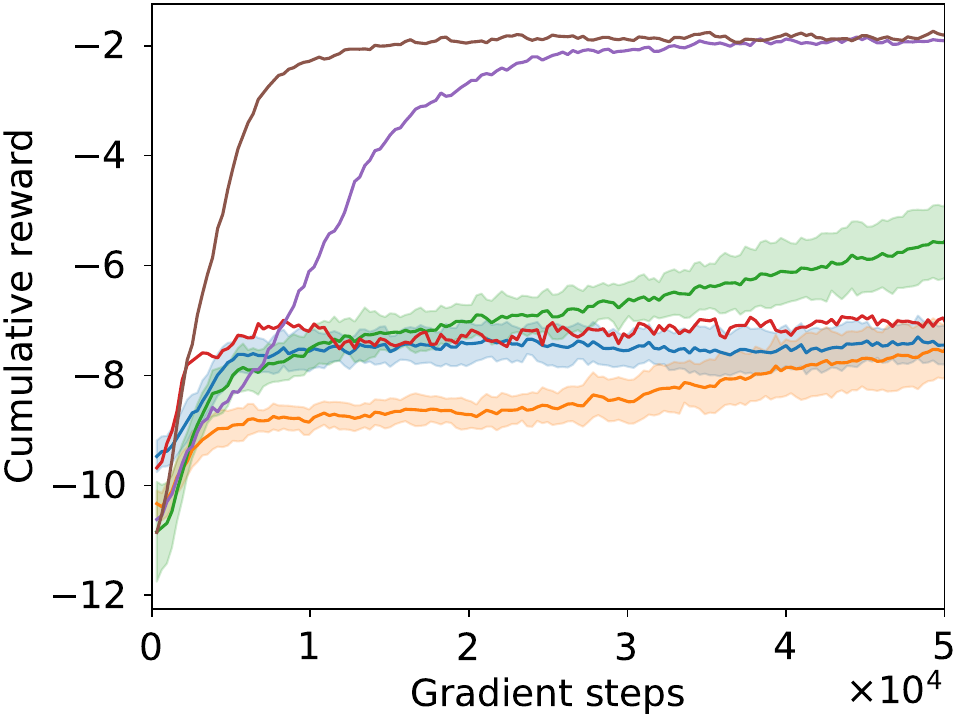}
    \caption{
        Learning curves for the Reacher task with gradient estimates of different accuracies.
        For each action space configuration, we compare the learning performance of an agent that uses accurate estimates of the policy gradient (estimated on batches of $10^5$ samples) to the learning performance of the default agent that uses rough estimates of the policy gradient (estimated on batches of $64$ samples).
        Note that the x-axis uses gradient steps instead of environment steps to account for the fact that more data is collected for the accurate gradient estimates.
        More accurate gradients improve the convergence rate and final performance significantly for the position and velocity control configurations but not for the torque control configuration.
    }
    \label{fig:gym_reacher_learning_curves_different_batch_sizes}
\end{figure}

In this section, we explore the effect of training with more accurate gradient estimates.
Instead of the batches of 64 samples that are commonly used in PPO to predict the policy gradient, we utilize $100{,}000$ samples to get significantly more accurate gradient estimates.
The learning curves in \cref{fig:gym_reacher_learning_curves_different_batch_sizes} indicate that more accurate gradient estimates can significantly accelerate learning and improve the final performance in certain cases.
However, the torque control configuration still performs poorly, indicating that the action representation influences factors of the learning process beyond the quality of the gradient estimates.

\subsection{Defining states and actions in the same space}
\label{app:states_and_actions_in_the_same_space}

In this section, we explore the interplay between state and action representations.
Particularly, we investigate whether a task becomes easier to learn if both the states and actions are defined in the same space.
Intuitively, such a setting should make it easier for the agent to understand the relationship between actions and resulting states.

We test this hypothesis on the Reacher task.
To that end, we define a modified version of the task that defines  the target position in joint space instead of end-effector space.
Modifying the state space of the original Reacher task, we arrive at the following definition of the state $\state$
\begin{equation}
    \state = \left(\pos, \pos_\mathrm{target}, \vel, \Delta \pos_\mathrm{target} \right),
\end{equation}
where $\pos$ denotes the current joint angles of the manipulator, $\pos_\mathrm{target}$ the target position expressed in joint angles, $\vel$ the current joint velocities, and $\Delta\pos_\mathrm{target}$ the joint angle distance to the target.

To ensure consistency, we also define the reward in joint space\footnote{
With the reward from the original task, which is defined in end-effector space, two configurations solve the task. For clarity, we would need to include both configurations in the observations.
}.
Specifically, we use the norm of the joint differences as the reward function.
\begin{equation}
    r(\state, \act{}) = \left\lVert \frac{2}{\pi}\,\Delta\pos_\mathrm{target} \right\rVert \qquad\text{with}~\state = \left(\pos, \pos_\mathrm{target}, \vel, \Delta \pos_\mathrm{target} \right)
\end{equation}
The factor $\frac{2}{\pi}$ ensures that the reward is in the same range as the end-effector-space reward of the original task.

\begin{figure}[H]
    \centering
    \vspace{\baselineskip}
    \includegraphics[width=0.5\textwidth]{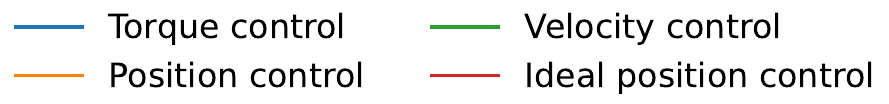} \\
    \vspace{0.2cm}
    \includegraphics[width=0.55\textwidth]{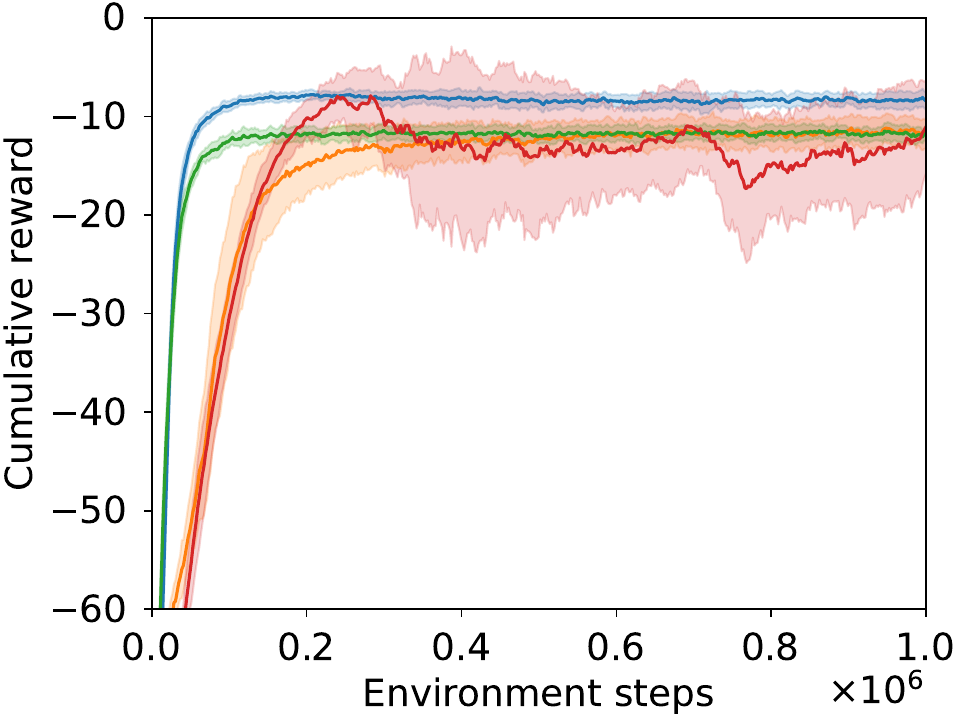}
    \caption{Learning curves for the modified Reacher task, where the target states are expressed in joint space and the reward is defined as the difference to the target joint angles.
    Although the task is intuitively very simple in the position control configurations, the torque control configuration achieves superior performance and has lower variance across random seeds.}
    \label{fig:learning_curves_gym_reacher_joint_space}
\end{figure}

On this task, we execute PPO with the three action spaces defined in \cref{sec:action_spaces}.
To assess the effects of the simple linear controller not always reaching the target position perfectly, we introduce the \emph{ideal position control} configuration in addition.
In this configuration, we remove the linear controller and simply set the simulation state to the target joint angles at each time step.
This scheme emulates a position controller that always reaches the given target positions perfectly in a single step.
Note that the task is quite simple in this configuration since the optimal policy is to simply take the target position from the state and apply it as action at every time step.
The learning curves of PPO with the different action spaces on the joint space Reacher task are given in \cref{fig:learning_curves_gym_reacher_joint_space}.

Surprisingly, torque control outperforms both position control configurations, even though the task is intuitively quite easy to solve in these configurations.
This result invalidates the hypothesis that defining the state and action in the same space is always superior to using different representations.
Moreover, this finding further illustrates that it is often challenging to get an intuitive understanding of the reasons for task difficulty in reinforcement learning.

\end{document}